\documentclass[10pt,twocolumn,letterpaper]{article}

\usepackage{cvpr}              

\usepackage{graphicx}
\usepackage{amsmath}
\usepackage{amssymb}
\usepackage{booktabs}
\usepackage{xspace}
\usepackage{multirow}
\usepackage{array}

\usepackage[pagebackref,breaklinks,colorlinks]{hyperref}

\usepackage[capitalize]{cleveref}
\crefname{section}{Sec.}{Secs.}
\Crefname{section}{Section}{Sections}
\Crefname{table}{Table}{Tables}
\crefname{table}{Tab.}{Tabs.}

\def\cvprPaperID{*****} 
\def\confName{CVPR}
\def\confYear{2022}

\newcommand{\methodname}{ConsistNet\xspace}

\newcommand{\methodsymbol}{\mathcal{M}}
\newcommand{\zero}{Zero123\xspace}

\newcommand{\vx}{\mathbf{x}}

\newcommand{\specialcell}[2][c]{%
  \begin{tabular}[#1]{@{}c@{}}#2\end{tabular}
}
  
\begin{document}
\catcode`\#=12

\twocolumn[{
\begin{@twocolumnfalse}

\title{\methodname: Enforcing 3D Consistency for Multi-view Images Diffusion}

\author{Jiayu Yang$^{1,2}$, \;\;Ziang Cheng$^{1,2}$, \;\;Yunfei Duan$^{1}$, \;\;Pan Ji$^{1}$, \;\; Hongdong Li$^{2}$\\
$^1$Tencent, $^2$Australian National University \\ \tt\small \{jiayu.yang, ziang.cheng, hongdong.li\}@anu.edu.au, \\ \tt\small kownseduan@global.tencent.com, panji@tencent.com} 

\maketitle

\thispagestyle{empty}

\vspace{-0.5cm}
\begin{center}
\setlength\tabcolsep{0pt}
\small
\renewcommand{\arraystretch}{0}
\begin{tabular}{c|cccccccc}
\includegraphics[trim=60 60 60 60, clip, width=0.11\linewidth]{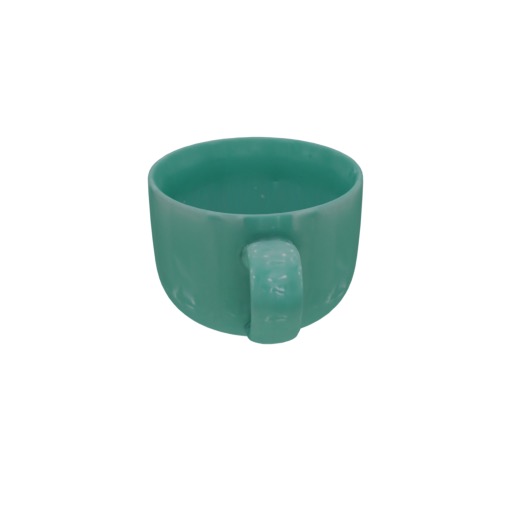} &
\includegraphics[trim=25 25 25 25, clip, width=0.11\linewidth]{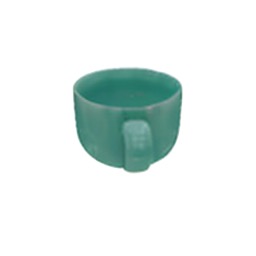} &
\includegraphics[trim=25 25 25 25, clip, width=0.11\linewidth]{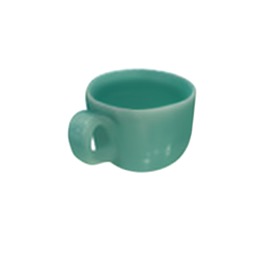} &
\includegraphics[trim=25 25 25 25, clip, width=0.11\linewidth]{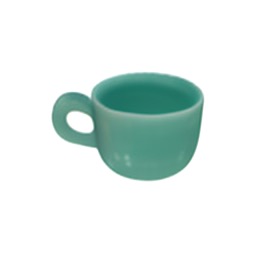} &
\includegraphics[trim=25 25 25 25, clip, width=0.11\linewidth]{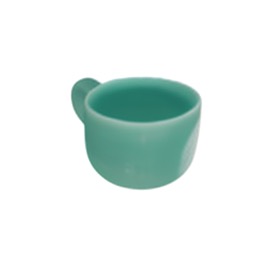} &
\includegraphics[trim=25 25 25 25, clip, width=0.11\linewidth]{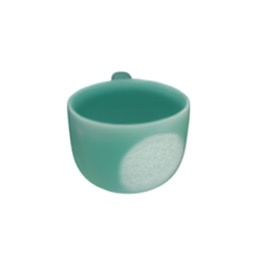} &
\includegraphics[trim=25 25 25 25, clip, width=0.11\linewidth]{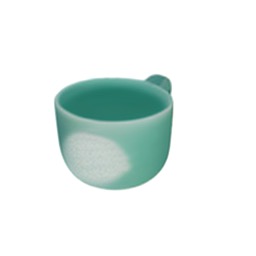} &
\includegraphics[trim=25 25 25 25, clip, width=0.11\linewidth]{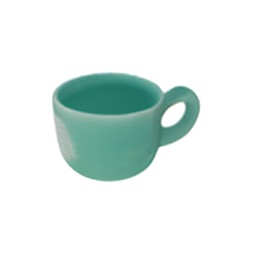} &
\includegraphics[trim=25 25 25 25, clip, width=0.11\linewidth]{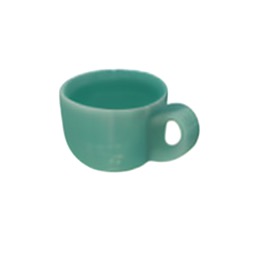} \\

\includegraphics[trim=60 60 60 60, clip, width=0.11\linewidth]{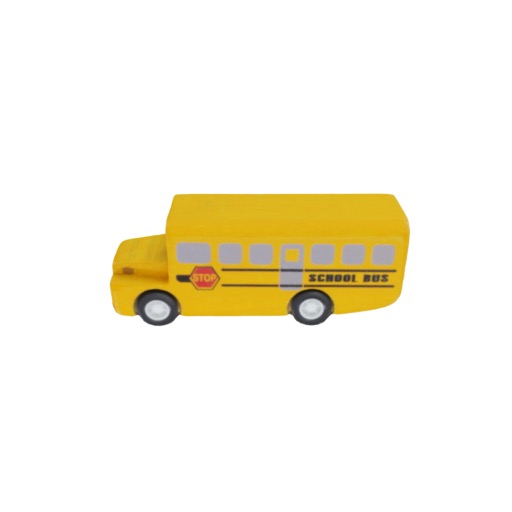} &
\includegraphics[trim=25 25 25 25, clip, width=0.11\linewidth]{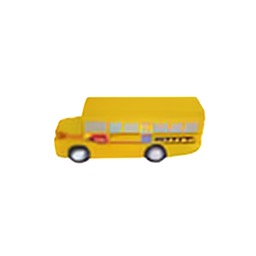} &
\includegraphics[trim=25 25 25 25, clip, width=0.11\linewidth]{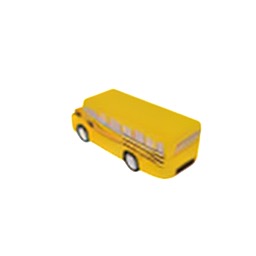} &
\includegraphics[trim=25 25 25 25, clip, width=0.11\linewidth]{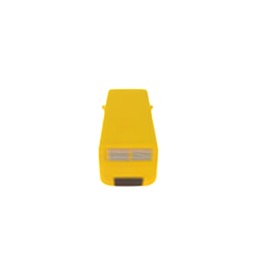} &
\includegraphics[trim=25 25 25 25, clip, width=0.11\linewidth]{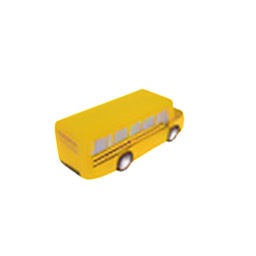} &
\includegraphics[trim=25 25 25 25, clip, width=0.11\linewidth]{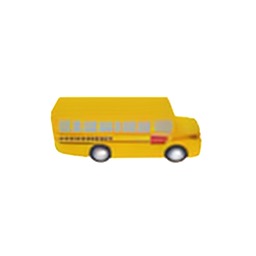} &
\includegraphics[trim=25 25 25 25, clip, width=0.11\linewidth]{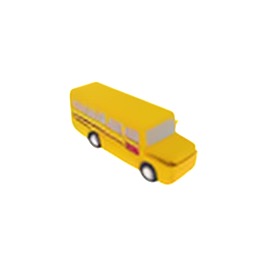} &
\includegraphics[trim=25 25 25 25, clip, width=0.11\linewidth]{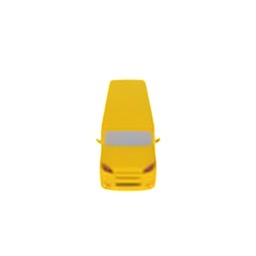} &
\includegraphics[trim=25 25 25 25, clip, width=0.11\linewidth]{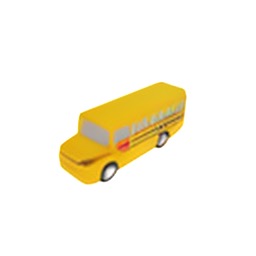} \\

\includegraphics[trim=60 60 60 60, clip, width=0.11\linewidth]{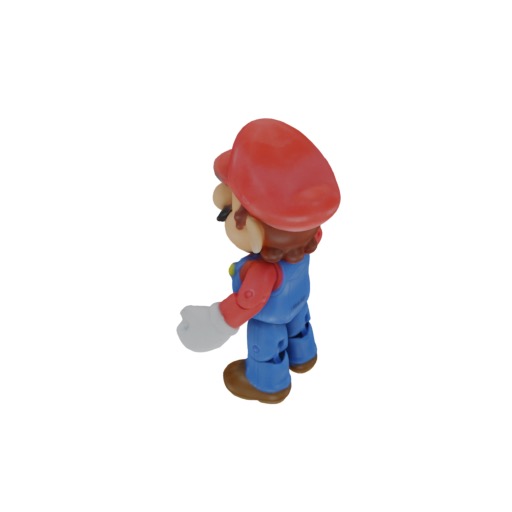} &
\includegraphics[trim=25 25 25 25, clip, width=0.11\linewidth]{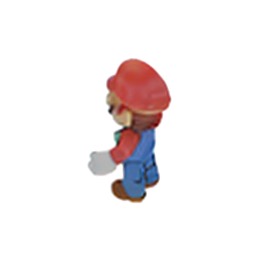} &
\includegraphics[trim=25 25 25 25, clip, width=0.11\linewidth]{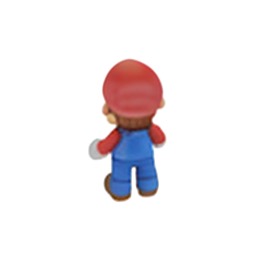} &
\includegraphics[trim=25 25 25 25, clip, width=0.11\linewidth]{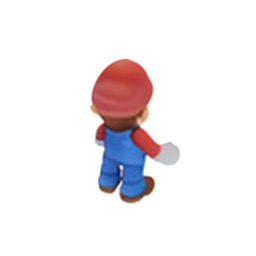} &
\includegraphics[trim=25 25 25 25, clip, width=0.11\linewidth]{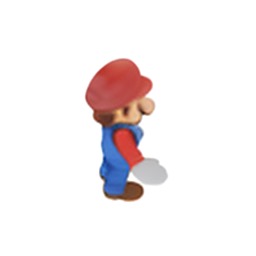} &
\includegraphics[trim=25 25 25 25, clip, width=0.11\linewidth]{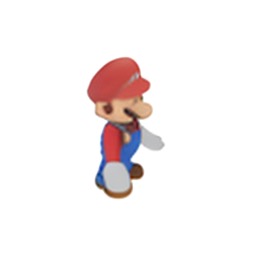} &
\includegraphics[trim=25 25 25 25, clip, width=0.11\linewidth]{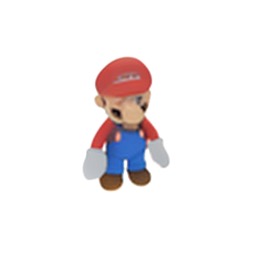} &
\includegraphics[trim=25 25 25 25, clip, width=0.11\linewidth]{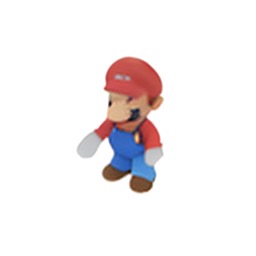} &
\includegraphics[trim=25 25 25 25, clip, width=0.11\linewidth]{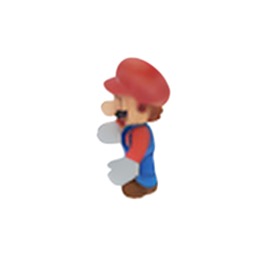} \\

\includegraphics[trim=60 60 60 60, clip, width=0.11\linewidth]{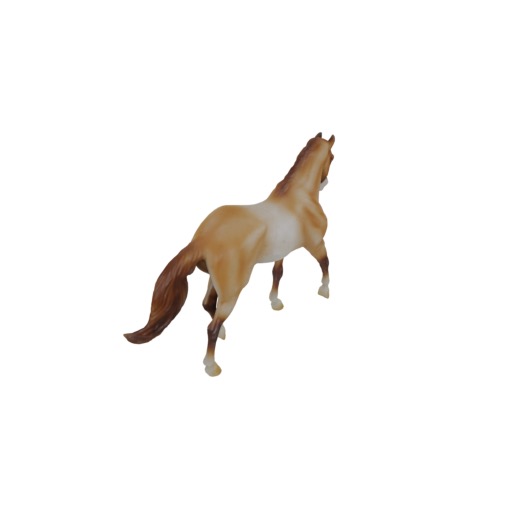} &
\includegraphics[trim=25 25 25 25, clip, width=0.11\linewidth]{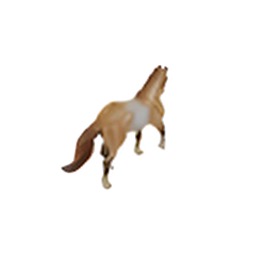} &
\includegraphics[trim=25 25 25 25, clip, width=0.11\linewidth]{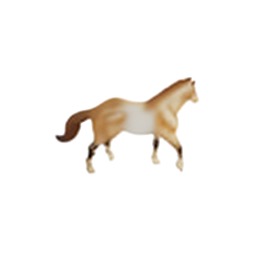} &
\includegraphics[trim=25 25 25 25, clip, width=0.11\linewidth]{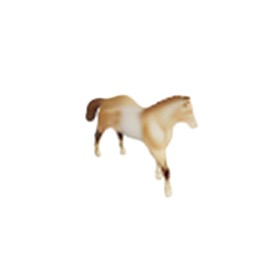} &
\includegraphics[trim=25 25 25 25, clip, width=0.11\linewidth]{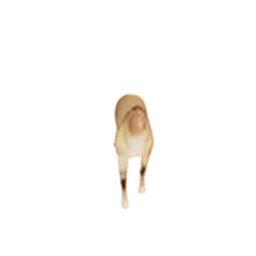} &
\includegraphics[trim=25 25 25 25, clip, width=0.11\linewidth]{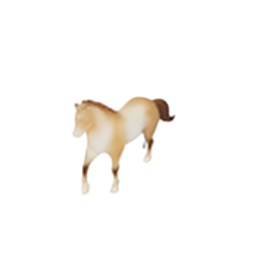} &
\includegraphics[trim=25 25 25 25, clip, width=0.11\linewidth]{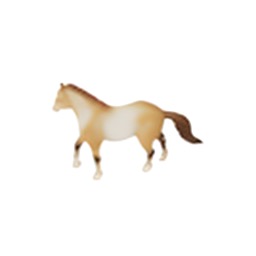} &
\includegraphics[trim=25 25 25 25, clip, width=0.11\linewidth]{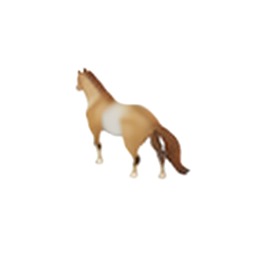} &
\includegraphics[trim=25 25 25 25, clip, width=0.11\linewidth]{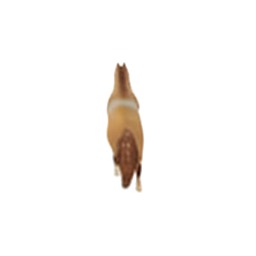} \\

\includegraphics[trim=60 60 60 60, clip, width=0.11\linewidth]{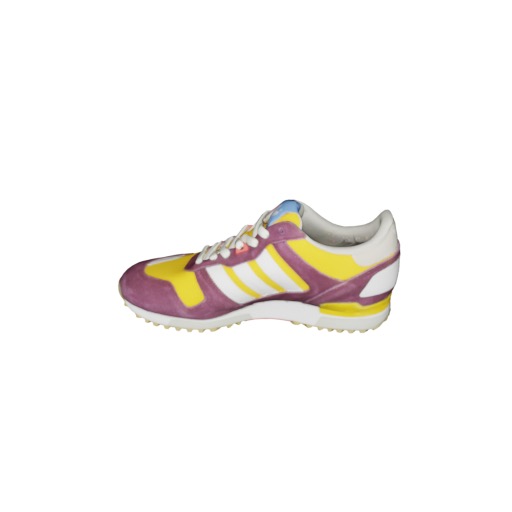} &
\includegraphics[trim=25 25 25 25, clip, width=0.11\linewidth]{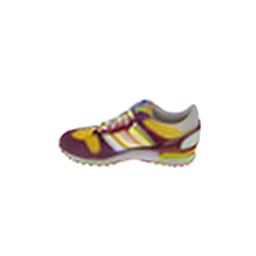} &
\includegraphics[trim=25 25 25 25, clip, width=0.11\linewidth]{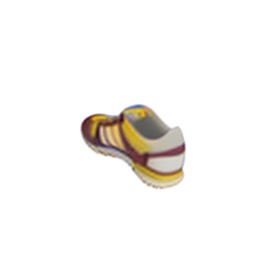} &
\includegraphics[trim=25 25 25 25, clip, width=0.11\linewidth]{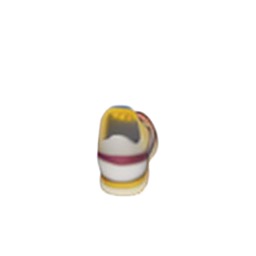} &
\includegraphics[trim=25 25 25 25, clip, width=0.11\linewidth]{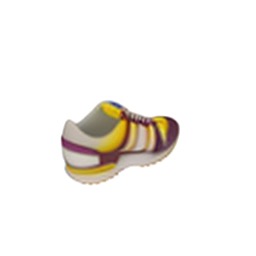} &
\includegraphics[trim=25 25 25 25, clip, width=0.11\linewidth]{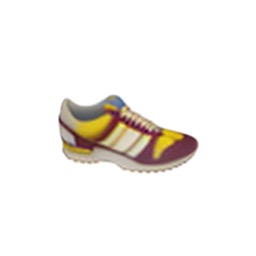} &
\includegraphics[trim=25 25 25 25, clip, width=0.11\linewidth]{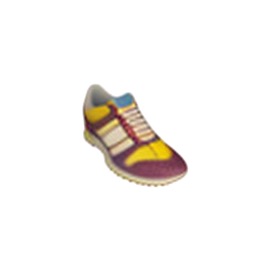} &
\includegraphics[trim=25 25 25 25, clip, width=0.11\linewidth]{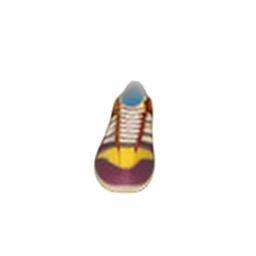} &
\includegraphics[trim=25 25 25 25, clip, width=0.11\linewidth]{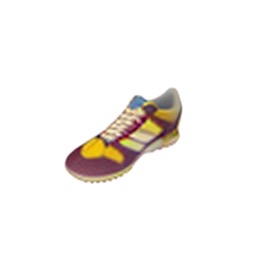} \\

\includegraphics[trim=60 60 60 60, clip, width=0.11\linewidth]{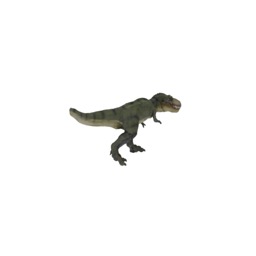} &
\includegraphics[trim=60 60 60 60, clip, width=0.11\linewidth]{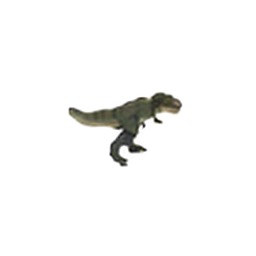} &
\includegraphics[trim=60 60 60 60, clip, width=0.11\linewidth]{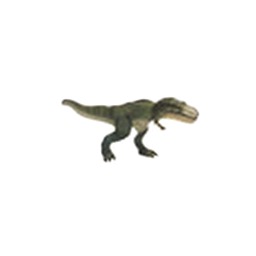} &
\includegraphics[trim=60 60 60 60, clip, width=0.11\linewidth]{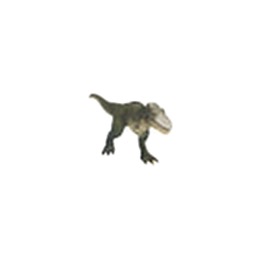} &
\includegraphics[trim=60 60 60 60, clip, width=0.11\linewidth]{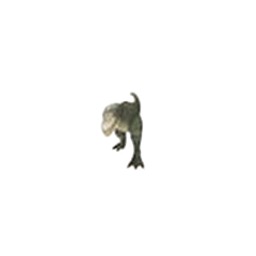} &
\includegraphics[trim=60 60 60 60, clip, width=0.11\linewidth]{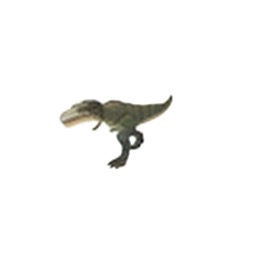} &
\includegraphics[trim=60 60 60 60, clip, width=0.11\linewidth]{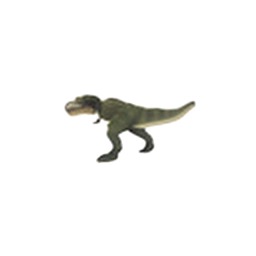} &
\includegraphics[trim=60 60 60 60, clip, width=0.11\linewidth]{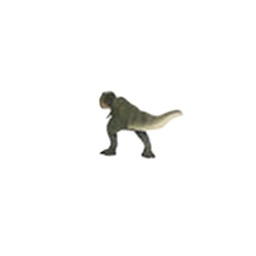} &
\includegraphics[trim=60 60 60 60, clip, width=0.11\linewidth]{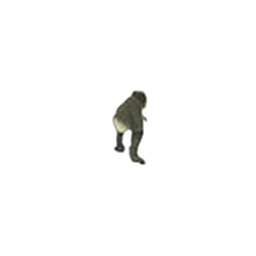} \\

\includegraphics[trim=10 10 10 10, clip, width=0.11\linewidth]{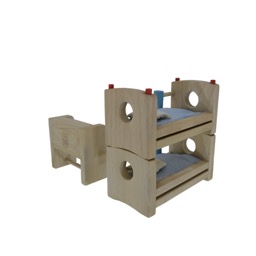} &
\includegraphics[trim=10 10 10 10, clip, width=0.11\linewidth]{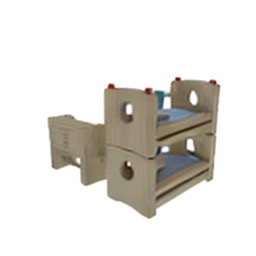} &
\includegraphics[trim=10 10 10 10, clip, width=0.11\linewidth]{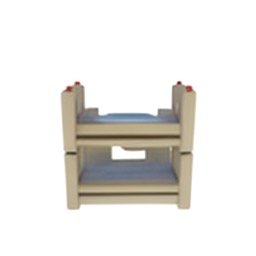} &
\includegraphics[trim=10 10 10 10, clip, width=0.11\linewidth]{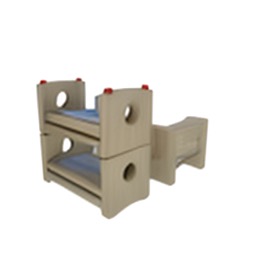} &
\includegraphics[trim=10 10 10 10, clip, width=0.11\linewidth]{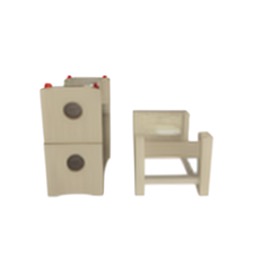} &
\includegraphics[trim=10 10 10 10, clip, width=0.11\linewidth]{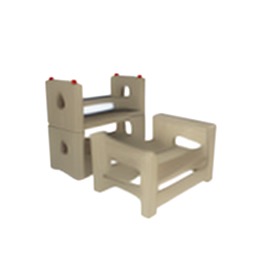} &
\includegraphics[trim=10 10 10 10, clip, width=0.11\linewidth]{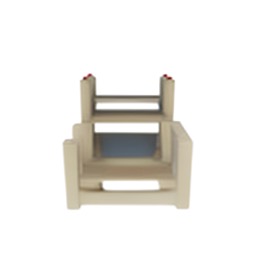} &
\includegraphics[trim=10 10 10 10, clip, width=0.11\linewidth]{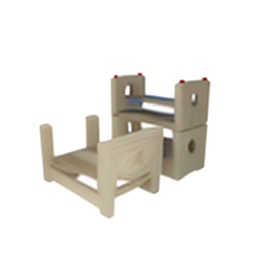} &
\includegraphics[trim=10 10 10 10, clip, width=0.11\linewidth]{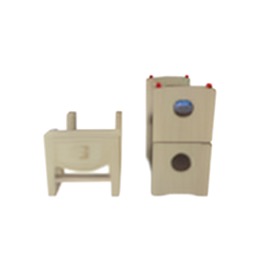} \\

\includegraphics[trim=24 24 24 24, clip, width=0.11\linewidth]{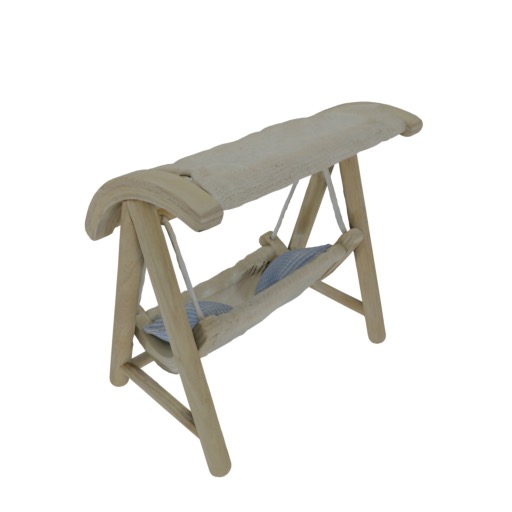} &
\includegraphics[trim=10 10 10 10, clip, width=0.11\linewidth]{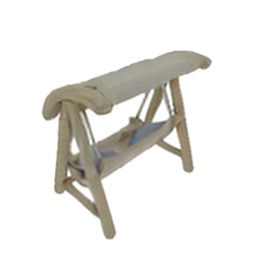} &
\includegraphics[trim=10 10 10 10, clip, width=0.11\linewidth]{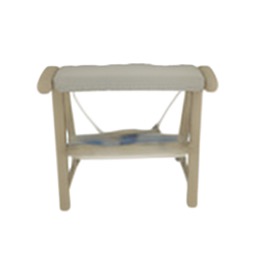} &
\includegraphics[trim=10 10 10 10, clip, width=0.11\linewidth]{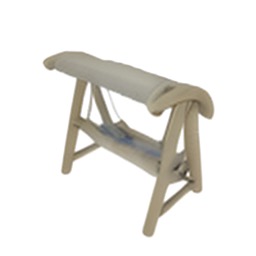} &
\includegraphics[trim=10 10 10 10, clip, width=0.11\linewidth]{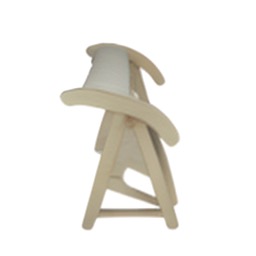} &
\includegraphics[trim=10 10 10 10, clip, width=0.11\linewidth]{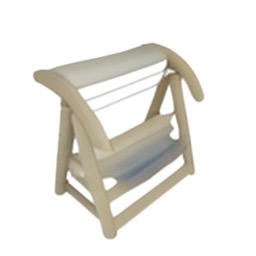} &
\includegraphics[trim=10 10 10 10, clip, width=0.11\linewidth]{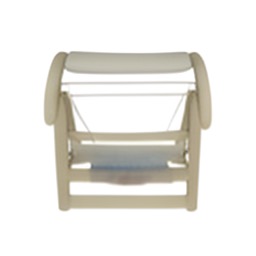} &
\includegraphics[trim=10 10 10 10, clip, width=0.11\linewidth]{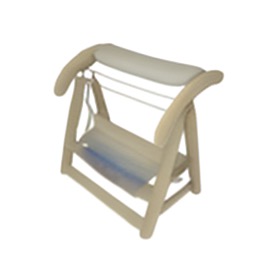} &
\includegraphics[trim=10 10 10 10, clip, width=0.11\linewidth]{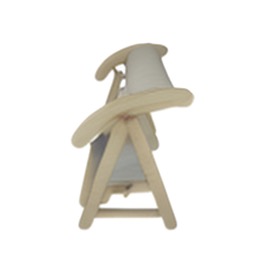} \\

Input image & \multicolumn{8}{c}{3D consistent images generated by \methodname + Zero123~\cite{zero123} }
           
\end{tabular}
\end{center}

\captionof{figure}{We present \methodname, a plug-in module for image diffusion models like Zero123~\cite{zero123} to generate multi-view consistent images. The \methodname is designed to be lightweight, allowing Zero123~\cite{zero123} to generate 16 multi-view consistent images in \textbf{40 seconds}. Animated visualizations can be seen at the project page \url{https://jiayuyang.github.io/Consist_Net}}
\label{fig:title}
\end{@twocolumnfalse}
}]
\clearpage

\begin{abstract}

Given a single image of a 3D object, this paper proposes a novel method (named \methodname) that is able to generate multiple images of the same object, as if seen they are captured from different viewpoints, while the 3D (multi-view) consistencies among those multiple generated images are effectively exploited.  Central to our method is a multi-view consistency block which enables information exchange across multiple single-view diffusion processes based on the underlying multi-view geometry principles. \methodname is an extension to the standard latent diffusion model, and consists of two sub-modules: (a) a view aggregation module that unprojects multi-view features into global 3D volumes and infer consistency, and (b) a ray aggregation module that samples and aggregate 3D consistent features back to each view to enforce consistency. Our approach departs from previous methods in multi-view image generation, in that it can be easily dropped-in pre-trained LDMs without requiring explicit pixel correspondences or depth prediction. Experiments show that our method effectively learns 3D consistency over a frozen \zero backbone and can generate 16 surrounding views of the object within 40 seconds on a single A100 GPU. Our code will be made available on \url{https://github.com/JiayuYANG/ConsistNet}.
\end{abstract}

\section{Introduction}
\label{sec:intro}

Recent advance of the Latent Diffusion Models (LDM)~\cite{rombach2022high,blattmann2023align} for image generation has brought about remarkable success in generating high-quality images with compelling details. However, when applied to generate multiple-view image of the same object, vanilla LDMs are unable to ensure the 3D-consistencies among the generated multiple images.  This is primarily due to there lacks mechanisms to enforce such 3D consistency information among the images taken at different viewpoints.

3D-consistent multi-view image diffusion models provide not only theoretical values, but hold major practical relevance, e.g. for 3D asset generation in VR/AR and video gaming applications.  Such diffusion models can either serve as a multi-view consistent image prior for 3D generation via the Score Distillation Sampling (SDS) loss~\cite{poole2022dreamfusion}, or allow direct reconstruction of 3D assets from once sampled images.

A recent work, \zero~\cite{zero123}, stands out as a promising approach for this purpose. It leverages the power of CLIP image embedding combined with camera embedding to produce semantically coherent images that are also viewpoint-aware.  However, such semantic multi-view consistent heuristics is rather weak, in the sense that the multiple-view images generated by \zero do not necessarily adhere to any shared 3D structure. In other words, the much desired multi-view geometry consistency is not explictly enforced in any effective manner. 

To address this challenge, we introduce a novel latent diffusion model. Instead of using a single diffusion model, we run multiple diffusion models in parallel, each dedicated to a specific viewpoint. We propose a plug-in multi-view consistency block, namely, \methodname.  This block ensures that the multiple images generated satisfy the underlying multi-view geometry principles (e.g., see Fig.~\ref{fig:title}).

In our method, the base diffusion models are pre-trained and remain frozen. The only trainable component is the \methodname block. This block can be plugged into every decoding layer of the denoising UNet. Its primary function is to gather multi-view feature maps and produce a residual feature map at every view point that reflects 3D consistency. This residual map is then added back into the corresponding decoder layers to enforce 3D consistency.

Our method, albeit based on the \zero backbone, surpasses it in terms of 3D consistency. Through extensive experiments, we have achieved marked improvement in 3D consistency, and our model exhibits commendable generality when exposed to unseen data.

\section{Related Work}
\label{sec:related_work}

\paragraph{Diffusion Model}
Denoising diffusion models have been applied to various tasks in computer vision, \eg, image enhancement \cite{whang2022deblurring,gao2023implicit,chen2023hierarchical}, style transfer and content editing \cite{ruiz2022dreambooth,zhang2023inversion,brooks2023instructpix2pix,liew2023magicedit,hertz2023delta}, image \cite{dhariwal2021diffusion,saharia2022photorealistic} video \cite{ho2022imagen,singer2022make,blattmann2023align,guo2023animatediff} and 3D shape generation. 

Classic diffusion models \cite{ho2020denoising,song2020score} generate images by reversing a Markov process where random noises are progressively added to clean images until the eventual distribution is Gaussian. Song~\etal~\cite{song2020denoising} propose DDIM sampling that uses an alternative non-Markovian formulation that significantly reduces number of denoising steps. A notable example of diffusion models is Latent diffusion models (LDM) \cite{rombach2022high}, where a variational autoencoder is first trained to compress natural images to a compact latent space where the diffusion process later takes place. 

There exist also methods for fine-tuning a pre-trained diffusion model, which allow the diffusion model to learn new concepts from a small dataset in an efficient manner  without the need to re-train the entire model  (\eg \cite{hu2021lora,ruiz2022dreambooth,gal2022image}).  ControlNet \cite{zhang2023adding} is yet another example which uses a zero-initialized ResNet block to piggyback a pre-trained diffusion model for enrich its expressive power.  ControlNet have demonstrated promising results on controlling single image generation with various forms of inputs, including depth map, normal map, sketch image and human pose. The role our new \methodname plays with respect to a pre-trained \zero is similar to what the ControlNet plays to a regular diffusion network. 

\paragraph{Sparse and Single View Novel View Synthesis}

A closely related task to multi-view consistent image generation is the task of Novel View Synthesis (NVS). Traditional NVS methods require a large number of input images from diverse viewpoints, and novel views are estimated from interpolating or extropolating those real input images (\eg \cite{gortler2023lumigraph,levoy2023light,mildenhall2021nerf, chen2021mvsnerf,shi2021self}. However, their performances is critically dependent on the viewpoint coverage and density of the input images. 

In contrast, recent generative models are capable of  'dreaming up' (\ie hallucinating) novel images that are from the input views' view-angles \cite{chan2023genvs,zou2023sparse3d,tseng2023consistent}. These methods may be able to maintain certain degree of geometry consistency between multiple views (by using \eg the epipolar constraint).  However, there is not guarantee that multi-view 3D consistencies among the generated novel views are well respected -- and this is the main task that this paper aims to solve.  Our method also draws inspiration from the cross-view attention mechanism used for novel view reconstruction \cite{wang2021ibrnet,wang2022attention,ren2023volrecon,chen2023matchnerf}. However, we use the idea in a novel way by un-projecting noisy features to 3D and re-project consistency information back to image to encourage 3D consistency. 

\paragraph{3D Consistent Image Generation} 
Several methods have been proposed for multi-view consistent image generation using diffusion models.  MVDream \cite{shi2023mvdream} fine-tunes a pre-trained image diffusion model on a multi-view dataset with a trainable attention module on the batch (multi-view) dimension. However, the attention module itself does not incorporate multi-view geometry, and the viewpoints are fixed. MVDiffusion \cite{tang2023mvdiffusion} achieves multi-view consistency by attending multi-view features with camera projection. However, it requires knowledge about scene geometry.

Concurrent to our work is SyncDreamer~\cite{liu2023syncdreamer}, which use a 3D-aware feature attention mechanism to synchronize features across views at every denoising step. It pre-process latent images from multiple views into a 3D volume and use the volume to guide U-Net to improve consistency. Differently, our \methodname block is designed to infer and improve 3D consistency operating within the U-Net itself acting as a plug-in module. Our method outperforms them both in speed and in geometry consistency on various elevation angles. 

\section{Method}
\label{sec:methodology}

\subsection{Latent Diffusion Model}
Latent Diffusion Model (LDM) is the backbone of our method. An LDM comprises two parts: a variational autoencoder (VAE) that compresses natural images into a computationally compact latent space $\mathcal{Z}$, and a denoising UNet that predicts the noise of a noisy latent representation. 

During training, random noises $\epsilon\sim\mathcal{N}$ are progressively added to $\vx_0\in\mathcal{Z}$ in a Markov chain of $t=1...N$ steps, and that $q(\vx_{t}|\vx_{t-1})$ is a Gaussian. The denoising UNet $\epsilon_\theta$ is trained to approximate the reverse process $q(\vx_{t-1}|\vx_t)$ by minimizing a lower bound loss term, in the form of
\begin{equation}
    L_t = \sum_{t=2}^{T} \mathbb{E}_{q} D_{KL}\big(q(\vx_{t-1}|\vx_0,\vx_t) \| p_\theta(\vx_{t-1}|\vx_t)\big). \label{eq:elob}
\end{equation}
The above minimization problem can be implemented as predicting the noise at each time step, leading to the following training loss
\begin{equation}
   L = \mathbb{E}_{\vx_0, \epsilon, t} \|\epsilon - \epsilon_\theta(\vx_t,t) \|_2^2 \label{eq:loss}.
\end{equation}

\begin{figure}[t]
  \centering
    \includegraphics[width=0.95\linewidth]{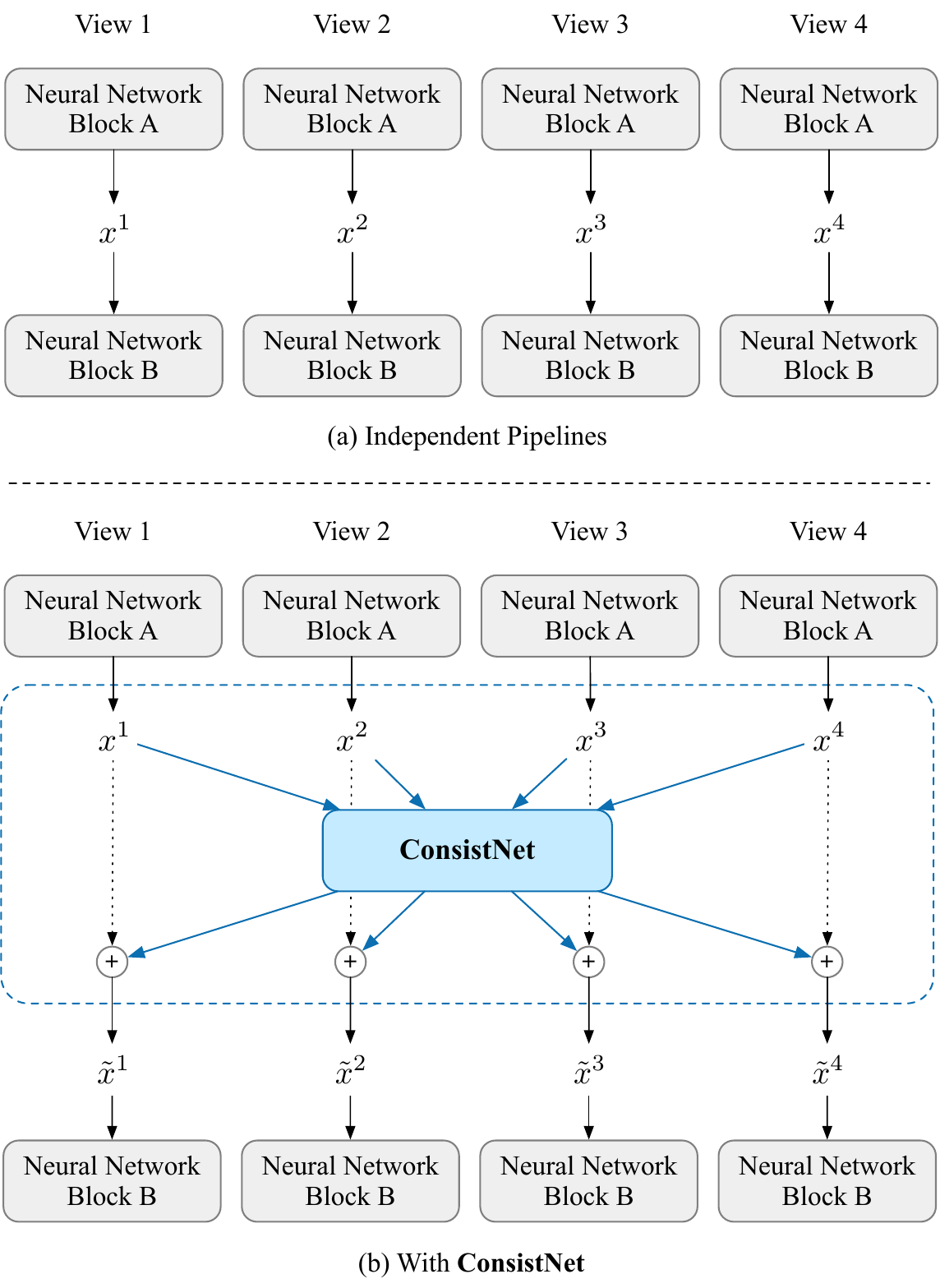}\\
  \captionof{figure}{Enforcing 3D consistency among individual network pipelines using \methodname. Trainable modules are marked in blue. (a) Network pipelines running independently do not have information over each other. (b) \methodname inter-connect pipelines and enforce 3D consistency. } 
  \label{fig:consistnet}
\end{figure}

Out goal is to extend the diffusion-based image generation process to the multi-view setting, where an independent random noise is progressively added to each multi-view image $\vx_t^1,...\vx_t^N$ during noising process. The resulted multi-view training loss is
\begin{equation}
   L = \mathbb{E}_{\vx_0^{1..N}, \epsilon, i, t} \|\epsilon - \epsilon^{1..N}_\theta(\vx_t^{1..N},t) \|_2^2 \label{eq:loss_mv}.
\end{equation}
Our multi-view diffusion model is realized by calling multiple single view LDMs in parallel.  To incorporate view consistency, we introduce a 3D aware plug-in module called \methodname block, that aggregates intermediate feature map through cross view projection.

\subsection{\methodname Block}
At the core of our method is an add-on block to pre-trained LDMs, called \methodname, that exchanges information between parallel LDMs running at different viewpoints based on underlying multi-view geometry principles, see Fig. \ref{fig:consistnet}. The \methodname block is trained end-to-end using the same loss function as defined in \eqref{eq:loss_mv}.

Denote by $i=1...N$ the viewpoint where every image is generated, the \methodname block $\methodsymbol$ is a residual attention block that connects all $N$ LDMs. It gathers a feature map $x^i_t$ from every LDM, and adds back to it a 3D-consistent feature map via residual connection,
\begin{equation}
    x^i_t \leftarrow x^i_t + \methodsymbol(i, \{x^j_t|j=1...N\}).
\end{equation}

\begin{figure}[t]
  \centering
    \includegraphics[width=0.96\linewidth]{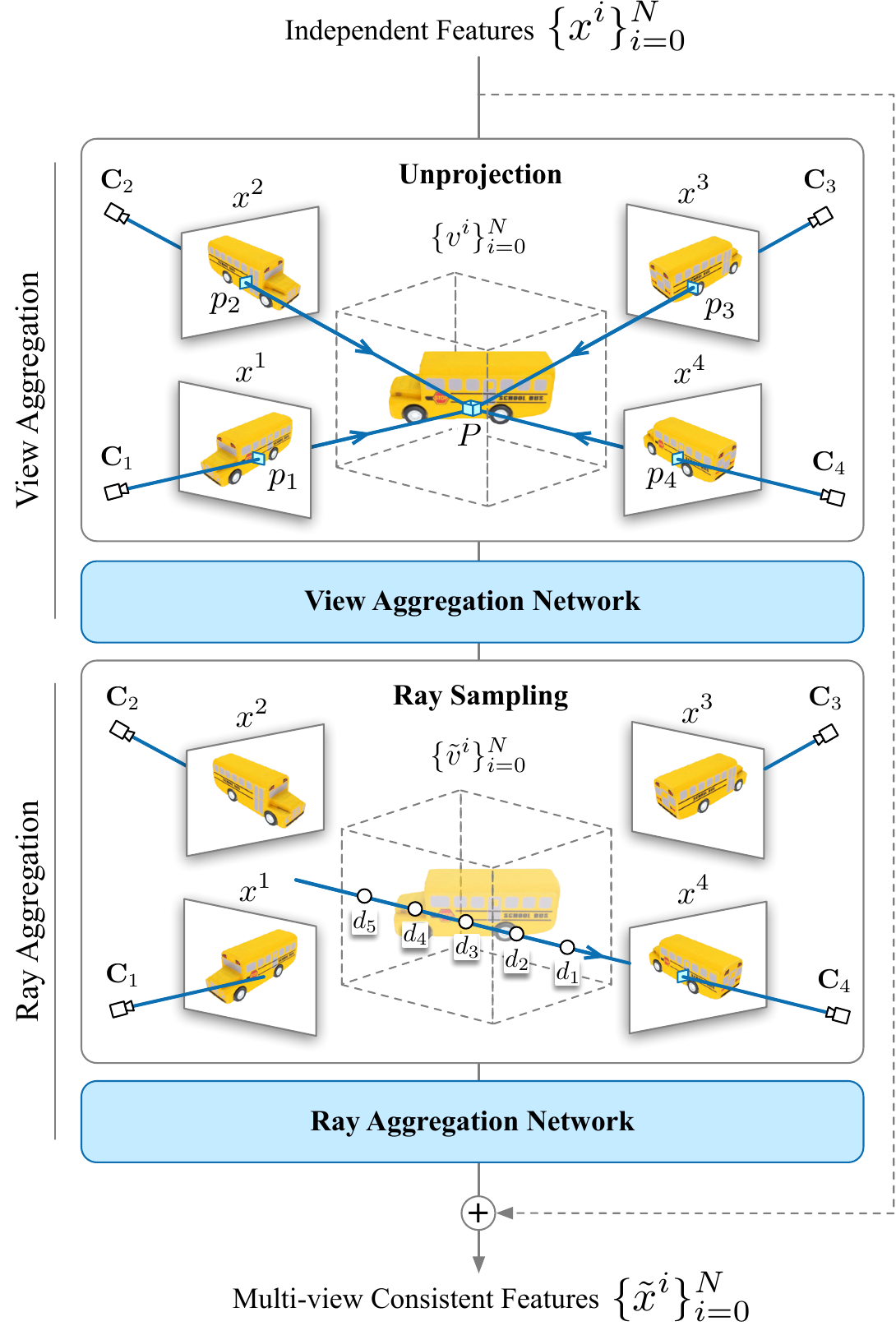}\\
  \captionof{figure}{\methodname Block. Our \methodname block consists of two sub-modules: (a) a view aggregation module that un-project image features to 3D and infer consistency by a view aggregation network, and (b) a ray aggregation module that sample 3D consistent features back to each view and use a ray aggregation network to enforce consistency. Trainable modules are marked in blue. } 
  \label{fig:block}
\end{figure}

The \methodname block comprises two sub-modules: (i) a view aggregation module that un-project feature maps $x^i_t$ into world feature volumes then use a view aggregation network to infer consistency, and (ii) a ray aggregation module that sample 3D consistent features back to each view and use a ray aggregation network to enforce consistency.

\paragraph{View Aggregation.} We start by unprojecting every feature maps $x^i$ onto a 3D volume in unified world coordinates, see Fig. \ref{fig:block}. This yields a volume $v^i$ for every viewpoint, defined as
\begin{equation}
    v^i = \Pi_i^{-1}(x^i) \oplus \text{PosEncode}(v^i_{cam}),
\end{equation}
where $\Pi^{-1}$ is the inverse camera projection by bi-linear interpolation, $\oplus$ denotes feature concatenation, and $v^i_{cam}$ is a fixed camera parameter volume that encodes the view direction and projection depth at each voxel. We use the sinusoidal position encoding for camera volume.

We use self-attention and 3D convolutions to implement the view aggregation network. All $N$ volumes are voxel-aligned, and are attended to each other by applying multi-headed self-attention layer in the $N$ views dimension. More formally,
\begin{equation}
    \bar{v}^i[P] = Attention(\{v^i[P]|i=1...N\}),
\end{equation}
where $P$ denotes voxels. We further process the volume $\bar{v}^i$ with few layers of 3D convolution. 

\paragraph{Ray Aggregation.} The volumes $\bar{v}^i$ gathers shared information between parallel multi-view LDMs, which is then delivered back to each LDM via a ray aggregation module, see Fig.~\ref{fig:block}. 

We first warp the world volumes $\bar{v}^i$ back to its corresponding camera frustum by uniformly sample depth along viewing ray of each pixel between minimum and maximum depth, use tri-linear interpolation to fetch the feature, and append the warped volume with its camera depth encoding for depth-wise attention.
\begin{equation}
    \tilde{v}^i = Attention\big(\text{Warp}_i(\bar{v}^i) \oplus \text{PosEncode}(d^i_{cam})\big).
\end{equation}

The warped volume is then projected to a 2D map $\tilde{x}^i$ by aggregating it along depth dimension. We implement this projection as the ray aggregation network, by a cross-attention layer using the feature map $x^i$ as query,
\begin{equation}
    \tilde{x}^i[r] = CrossAttention(x^i,\{\tilde{v}^i[d]|d\in[d_\text{near},d_\text{far}]\}),
\end{equation}
where $r$ is the ray traced back from camera $i$, and $d_\text{near}$ and $d_\text{far}$ are the near and far depth of camera frustum.

To allow fast training of \methodname on a pre-trained LDM, we add $\tilde{x}^i$ back to the original feature map ${x}^i$ via a residual layer whose weights are initialized to zero following ControlNet \cite{zhang2023adding}. The residual layer is implemented as an MLP that processes the feature map in a per-pixel manner without convolution. Hence the only trainable components are the two aggregation networks and the final MLP.

\section{\methodname for Multi-view Diffusion}
\begin{figure}[t]
  \centering
    \includegraphics[width=1.0\linewidth]{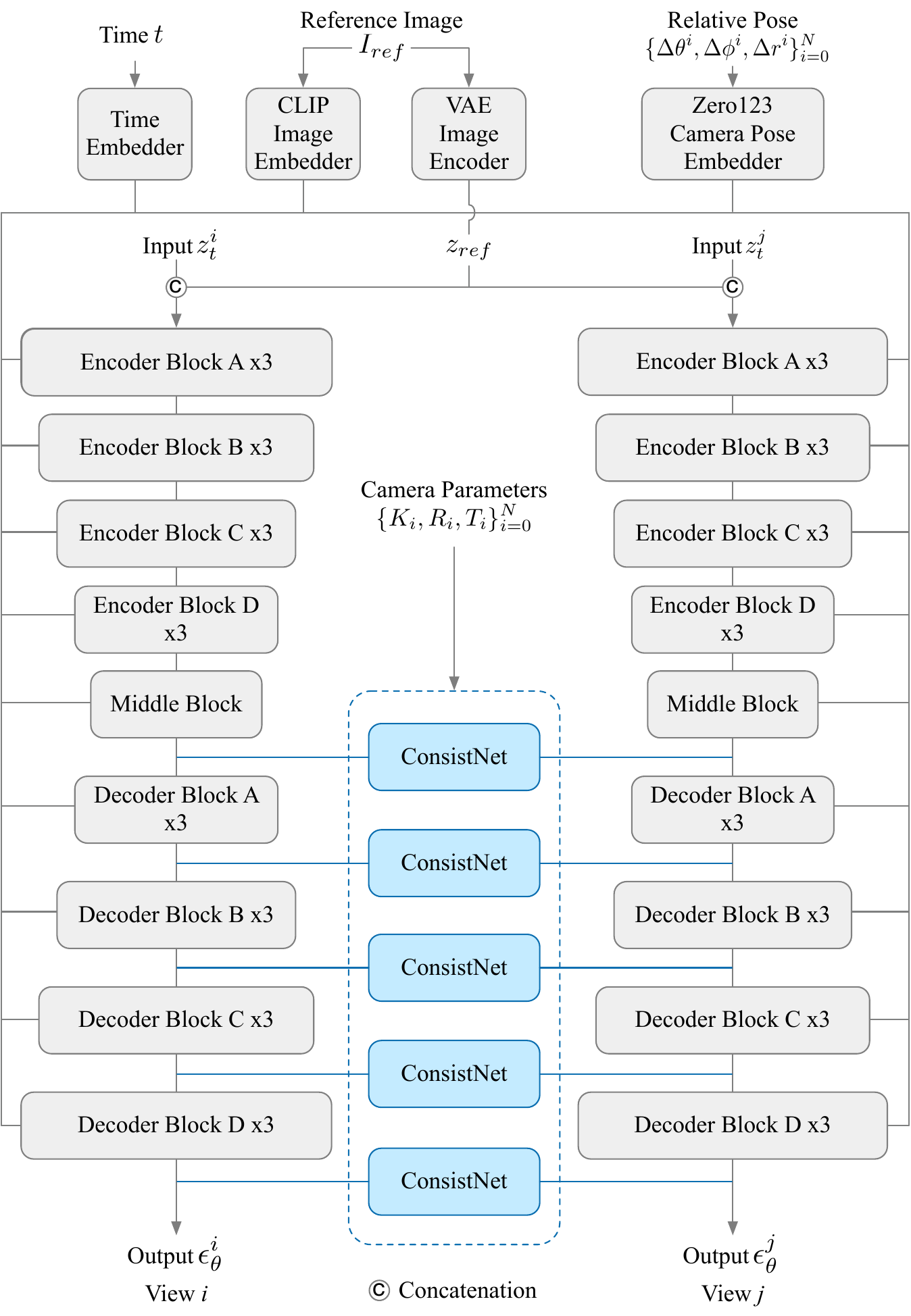}\\
  \captionof{figure}{\methodname plugged into the U-Net of Zero123~\cite{zero123}. Trainable modules are marked in blue. We plug \methodname block after every decoder block of \zero's U-Net to enforce 3D consistency.} 
  \label{fig:unet}
\end{figure}

We use \zero\cite{zero123} as backbone to showcase how \methodname can enforce 3D multi-view consistency in large pretrained LDM models. \zero is a specifically fine-tuned version of image variation LDM model, which follows the structure of Stable Diffusion~\cite{rombach2022high}, see Fig.~\ref{fig:unet}. The denoising U-Net~\cite{ronneberger2015unet} contains an encoder and a decoder, both of which have 12 blocks, and a mid block in between. Unlike Stable-Diffusion where the denoising U-Net is conditioned on the CLIP language embedding of text input, \zero is conditioned on the CLIP image embedding of the input image as well as the relative camera rotation from input view. Moreover, the reference image is encoded to a reference latent image and is concatenated with the noisy latent image as input to U-Net on every denoising step.

We instantiate several parallel pretrained \zero models, with each model dedicated to a specific viewpoint. Without the \methodname block, these diffusion processes would operate independently of each other, lacking the 3D consistency between shared views. To introduce 3D consistency, we insert a \methodname block to each decoder layer of the UNet, therefore allowing the feature maps to be mutually attended in a coarse to fine manner. An illustration is given in Fig \ref{fig:unet}.

\section{Experiments}
\label{sec:experiments}

\subsection{Datasets}
\noindent\textbf{Objaverse Dataset}~\cite{deitke2023objaverse} is a large-scale dataset containing 800K+ annotated 3D mesh objects.  We use this dataset for training and validation. We first filter out samples containing multiple objects by counting the number of bounding boxes in the scene. We render 16 images per object from uniformly distributed viewpoints surrounding the object. For each object, we randomly choose a positive elevation angle up to 30 degrees for all views. We render under a mixture of global lighting and point lighting from camera centre to produce desired shading effects (\eg normal direction and specular highlights) without introducing strong shadows. 

\noindent\textbf{Google Scanned Objects Dataset}~\cite{anthony2022google} is a open-source collection of over one thousand 3D-scanned household items. We use this entire dataset for evaluation only. We render objects from this dataset under the same setting as Objaverse dataset and manually set three options of elevation angles, 0, 15, and 30 degrees, to evaluate model performance on different elevation angles.

\subsection{Implementation Details}
Our \methodname blocks is plugged into a pre-trained and frozen \zero backbone. We train for 85k iterations with an AdamW optimizer \cite{loshchilov2017decoupled} and learning rate of $3\times 10^{-5}$. The training takes 46 hours on 8 A100 40G GPUs. We directly test our trained model on entire Google Scanned Objects \cite{anthony2022google} dataset without any fine-tuning. We use DDIM sampler~\cite{song2020denoising} with 50 denoising steps and generate 16 multi-view images for each object. We use a single A100 40GB GPU for all evaluations. We implement \methodname in HuggingFace Diffusers~\cite{von-platen-etal-2022-diffusers} framework.

\begin{table*}[]
\vspace{-0.1cm}
\centering
\resizebox{0.9\linewidth}{!}{
\begin{tabular}{c|l|ccccc|c}
\hline
Elevation           & Model                      & LPIPS\_Alex $\downarrow$ & LPIPS\_VGG $\downarrow$ & SSIM $\uparrow$ & MS-SSIM $\uparrow$ & PSNR $\uparrow$ & Runtime $\downarrow$    \\\hline\hline
\multirow{4}{*}{0}
                    & \zero~\cite{zero123}                   & 0.19    & 0.14    & 0.85 & 0.72        & 18.25 & 5s\\
                    & DreamFusion~\cite{poole2022dreamfusion}+\zero~\cite{zero123}               & 0.22    & 0.16   & 0.87 & 0.72        & 18.53  & 18min\\
                    & SyncDreamer~\cite{liu2023syncdreamer}                & 0.24    & 0.17   & 0.85 & 0.66        & 17.41  & 2min\\
                    & Ours & \textbf{0.15}    & \textbf{0.11}   & \textbf{0.89} & \textbf{0.82}        & \textbf{22.75}  & 40s \\\hline
\multirow{4}{*}{15} 
                    & \zero~\cite{zero123}                    & 0.23    & 0.17   & 0.83 & 0.65       & 16.59  & 5s \\
                    & DreamFusion~\cite{poole2022dreamfusion}+\zero~\cite{zero123}               & 0.23    & 0.16   & 0.85 & 0.69        & 17.76  & 18min \\
                    & SyncDreamer~\cite{liu2023syncdreamer}                 & 0.17    & 0.14   & 0.86 & 0.76        & 18.93  & 2min\\
                    & Ours & \textbf{0.12}    & \textbf{0.09}   & \textbf{0.90} & \textbf{0.86}        & \textbf{23.93}  & 40s \\\hline
\multirow{4}{*}{30} 
                    & \zero~\cite{zero123}                    & 0.27    & 0.18   & 0.83 & 0.61        & 16.10  & 5s \\
                    & DreamFusion~\cite{poole2022dreamfusion}+\zero~\cite{zero123}              & 0.14    & 0.12   & 0.88 & 0.84        & 21.53 & 18min \\
                    & SyncDreamer~\cite{liu2023syncdreamer}                & \textbf{0.10}    & \textbf{0.09}   & \textbf{0.90} & \textbf{0.88}        & \textbf{23.81}  & 2min\\
                    & Ours & 0.11    & \textbf{0.09}   & \textbf{0.90} & 0.86        & 23.67  & 40s\\
                    \hline
\end{tabular}
}
\vspace{-0.1cm}
\caption{\textbf{Google Scanned Objects Dataset.} Performance on different  elevation angle. Our model performs comparably to Syndreamer on elevation angle 30, but is able to generalize well on 0 and 15 degree elevations.}\label{tab:quantitative_results}
\end{table*}

\begin{figure*}[t]
\vspace{-0.3cm}
\begin{center}
\setlength\tabcolsep{1pt}
\small
\begin{tabular}{c|cccccccc}

\includegraphics[trim=60 60 60 60, clip, width=0.11\linewidth]{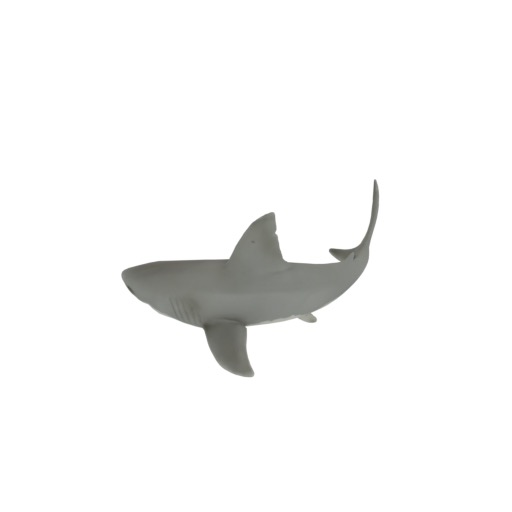} &
\includegraphics[trim=25 25 25 25, clip, width=0.11\linewidth]{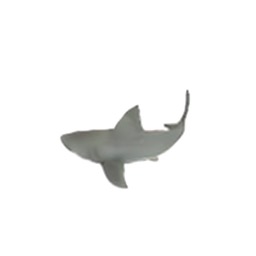} &
\includegraphics[trim=25 25 25 25, clip, width=0.11\linewidth]{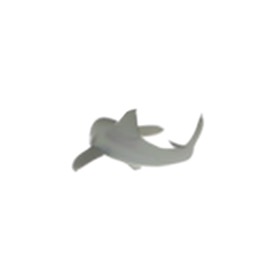} &
\includegraphics[trim=25 25 25 25, clip, width=0.11\linewidth]{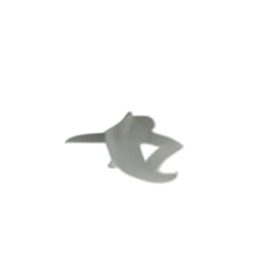} &
\includegraphics[trim=25 25 25 25, clip, width=0.11\linewidth]{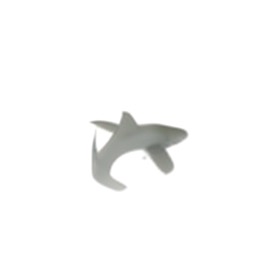} &
\includegraphics[trim=25 25 25 25, clip, width=0.11\linewidth]{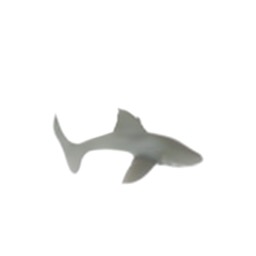} &
\includegraphics[trim=25 25 25 25, clip, width=0.11\linewidth]{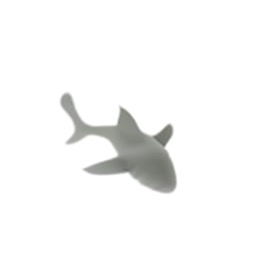} &
\includegraphics[trim=25 25 25 25, clip, width=0.11\linewidth]{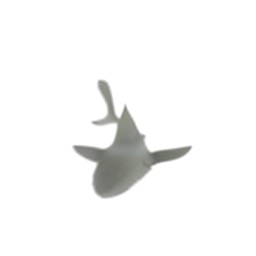} &
\includegraphics[trim=25 25 25 25, clip, width=0.11\linewidth]{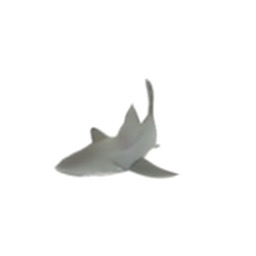} \\

\includegraphics[trim=60 60 60 60, clip, width=0.11\linewidth]{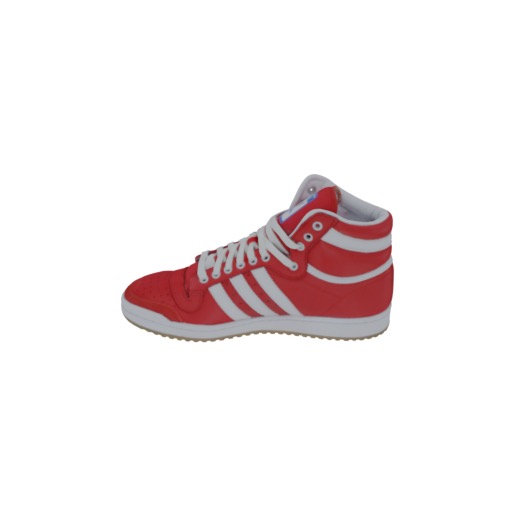} &
\includegraphics[trim=25 25 25 25, clip, width=0.11\linewidth]{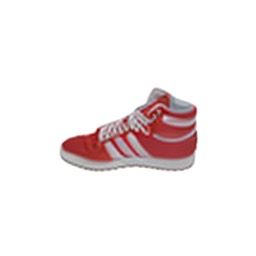} &
\includegraphics[trim=25 25 25 25, clip, width=0.11\linewidth]{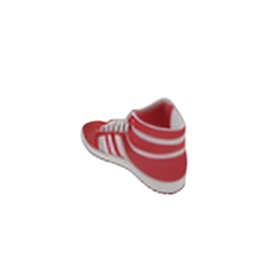} &
\includegraphics[trim=25 25 25 25, clip, width=0.11\linewidth]{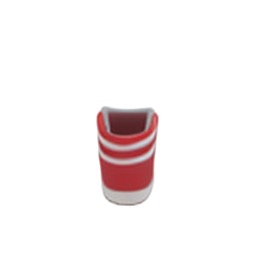} &
\includegraphics[trim=25 25 25 25, clip, width=0.11\linewidth]{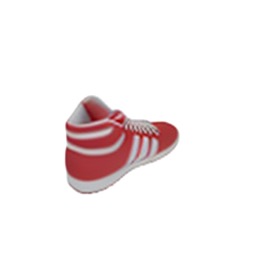} &
\includegraphics[trim=25 25 25 25, clip, width=0.11\linewidth]{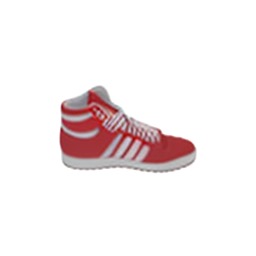} &
\includegraphics[trim=25 25 25 25, clip, width=0.11\linewidth]{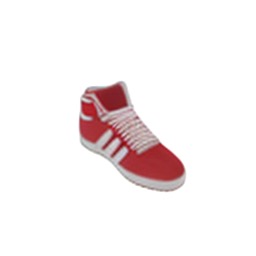} &
\includegraphics[trim=25 25 25 25, clip, width=0.11\linewidth]{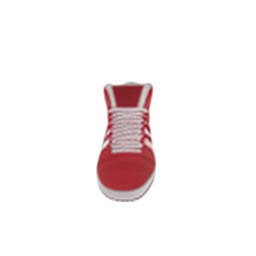} &
\includegraphics[trim=25 25 25 25, clip, width=0.11\linewidth]{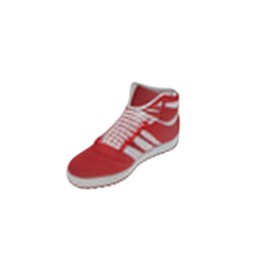} \\

\includegraphics[trim=60 60 60 60, clip, width=0.11\linewidth]{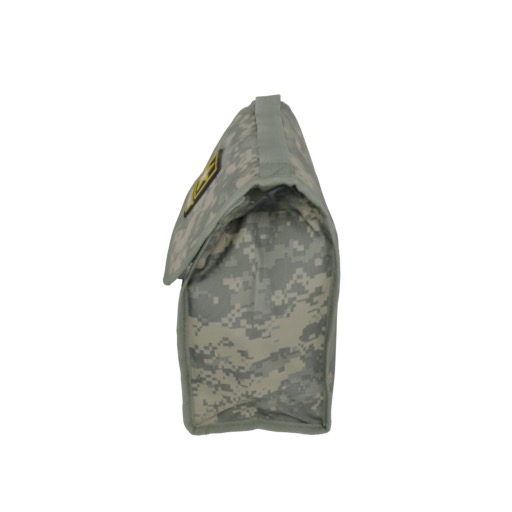} &
\includegraphics[trim=25 25 25 25, clip, width=0.11\linewidth]{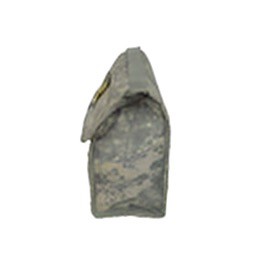} &
\includegraphics[trim=25 25 25 25, clip, width=0.11\linewidth]{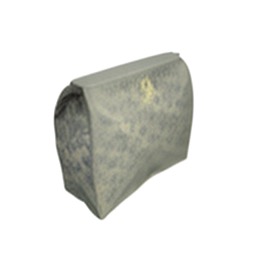} &
\includegraphics[trim=25 25 25 25, clip, width=0.11\linewidth]{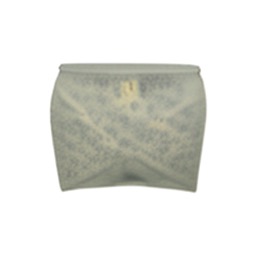} &
\includegraphics[trim=25 25 25 25, clip, width=0.11\linewidth]{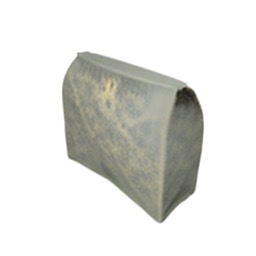} &
\includegraphics[trim=25 25 25 25, clip, width=0.11\linewidth]{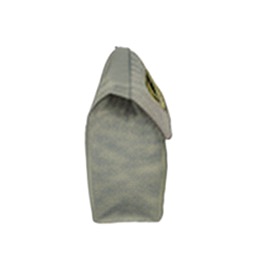} &
\includegraphics[trim=25 25 25 25, clip, width=0.11\linewidth]{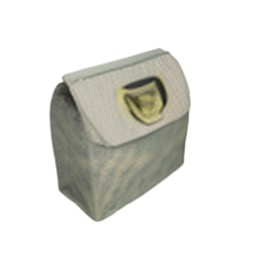} &
\includegraphics[trim=25 25 25 25, clip, width=0.11\linewidth]{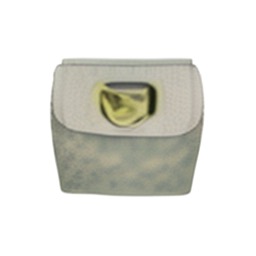} &
\includegraphics[trim=25 25 25 25, clip, width=0.11\linewidth]{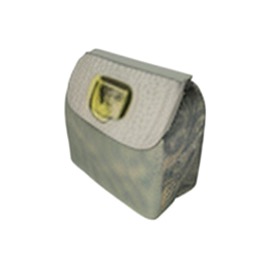} \\

\includegraphics[trim=60 60 60 60, clip, width=0.11\linewidth]{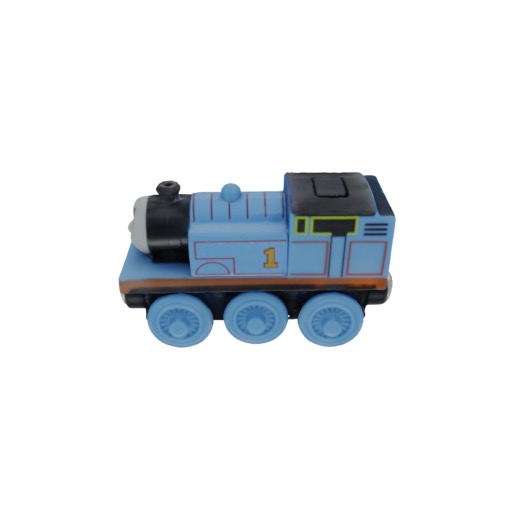} &
\includegraphics[trim=25 25 25 25, clip, width=0.11\linewidth]{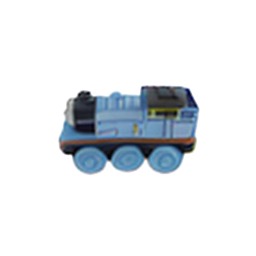} &
\includegraphics[trim=25 25 25 25, clip, width=0.11\linewidth]{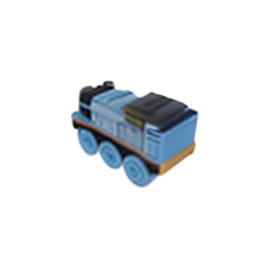} &
\includegraphics[trim=25 25 25 25, clip, width=0.11\linewidth]{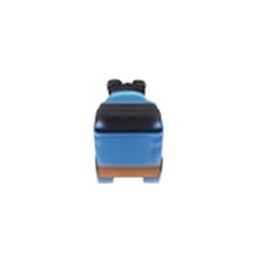} &
\includegraphics[trim=25 25 25 25, clip, width=0.11\linewidth]{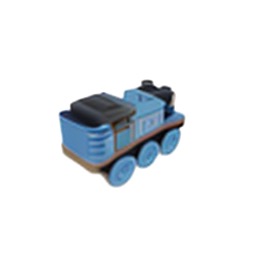} &
\includegraphics[trim=25 25 25 25, clip, width=0.11\linewidth]{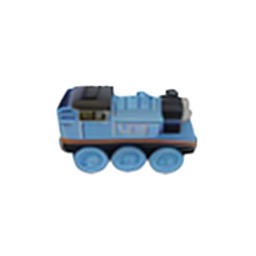} &
\includegraphics[trim=25 25 25 25, clip, width=0.11\linewidth]{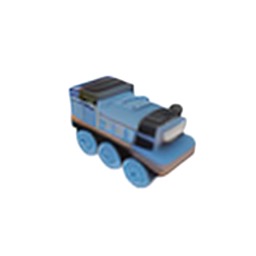} &
\includegraphics[trim=25 25 25 25, clip, width=0.11\linewidth]{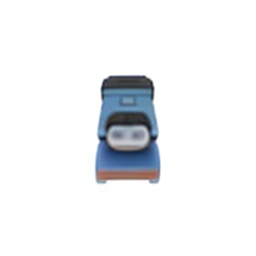} &
\includegraphics[trim=25 25 25 25, clip, width=0.11\linewidth]{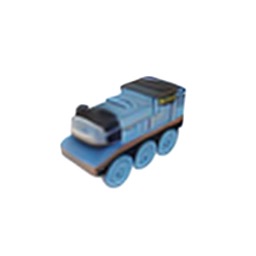} \\

\includegraphics[trim=60 60 60 60, clip, width=0.11\linewidth]{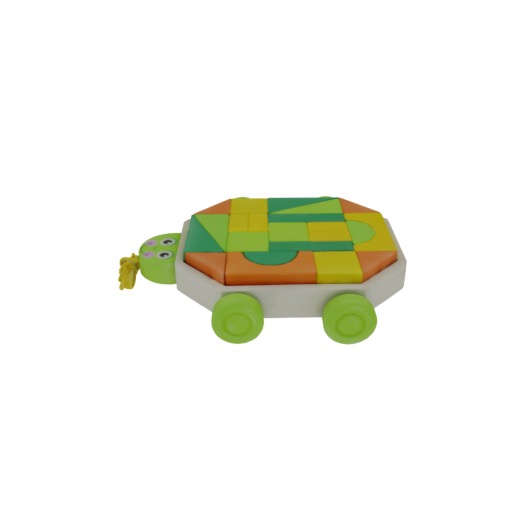} &
\includegraphics[trim=25 25 25 25, clip, width=0.11\linewidth]{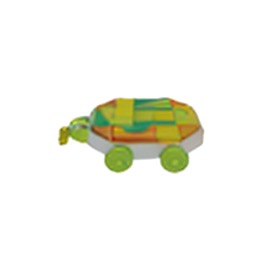} &
\includegraphics[trim=25 25 25 25, clip, width=0.11\linewidth]{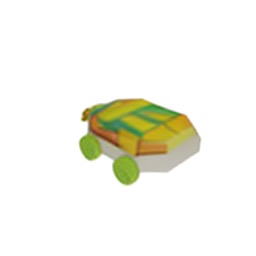} &
\includegraphics[trim=25 25 25 25, clip, width=0.11\linewidth]{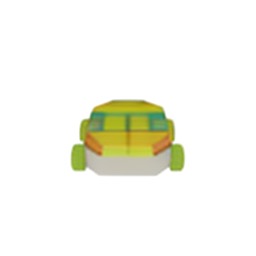} &
\includegraphics[trim=25 25 25 25, clip, width=0.11\linewidth]{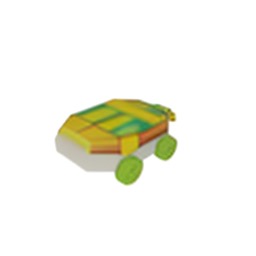} &
\includegraphics[trim=25 25 25 25, clip, width=0.11\linewidth]{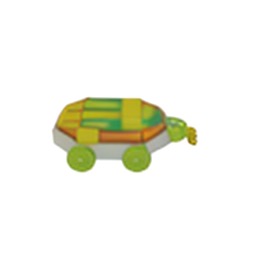} &
\includegraphics[trim=25 25 25 25, clip, width=0.11\linewidth]{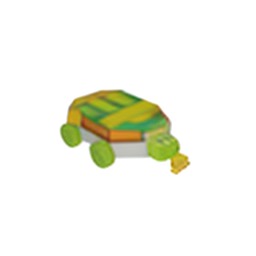} &
\includegraphics[trim=25 25 25 25, clip, width=0.11\linewidth]{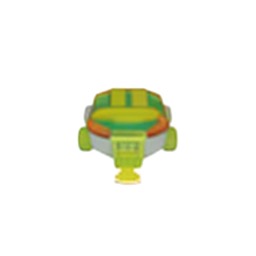} &
\includegraphics[trim=25 25 25 25, clip, width=0.11\linewidth]{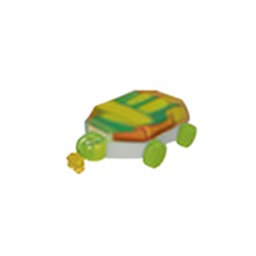} \\

\includegraphics[trim=0 0 0 0, clip, width=0.11\linewidth]{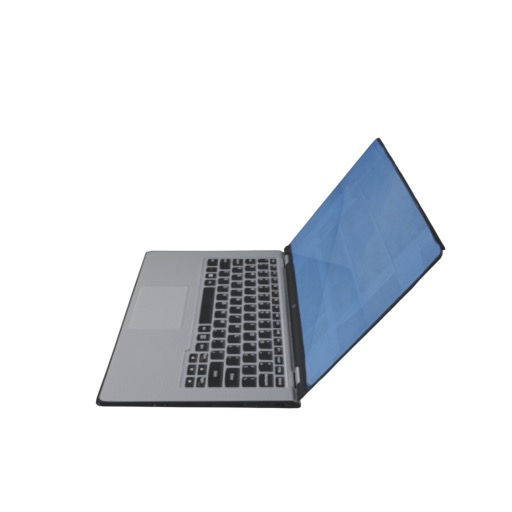} &
\includegraphics[trim=0 0 0 0, clip, width=0.11\linewidth]{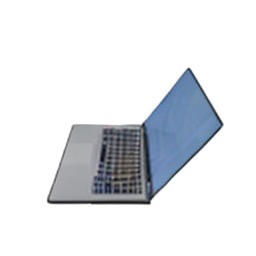} &
\includegraphics[trim=0 0 0 0, clip, width=0.11\linewidth]{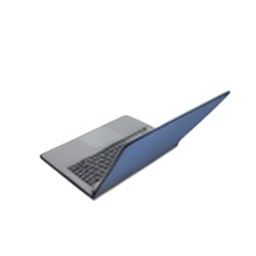} &
\includegraphics[trim=0 0 0 0, clip, width=0.11\linewidth]{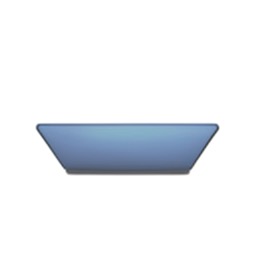} &
\includegraphics[trim=0 0 0 0, clip, width=0.11\linewidth]{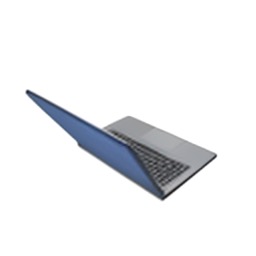} &
\includegraphics[trim=0 0 0 0, clip, width=0.11\linewidth]{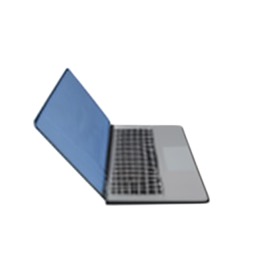} &
\includegraphics[trim=0 0 0 0, clip, width=0.11\linewidth]{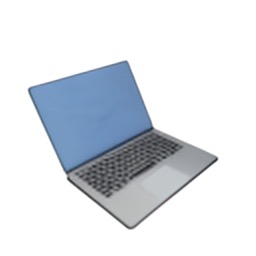} &
\includegraphics[trim=0 0 0 0, clip, width=0.11\linewidth]{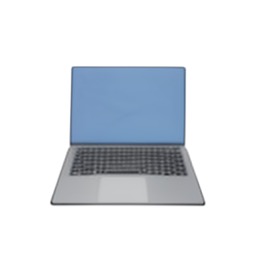} &
\includegraphics[trim=0 0 0 0, clip, width=0.11\linewidth]{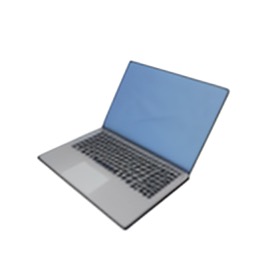} \\

\includegraphics[trim=40 40 40 40, clip, width=0.11\linewidth]{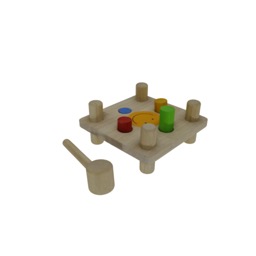} &
\includegraphics[trim=40 40 40 40, clip, width=0.11\linewidth]{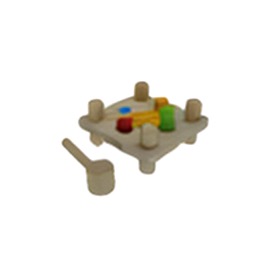} &
\includegraphics[trim=40 40 40 40, clip, width=0.11\linewidth]{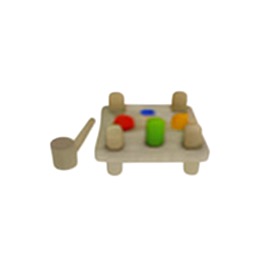} &
\includegraphics[trim=40 40 40 40, clip, width=0.11\linewidth]{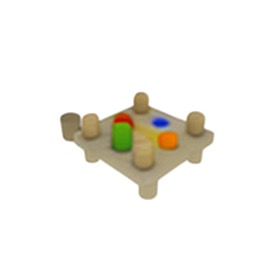} &
\includegraphics[trim=40 40 40 40, clip, width=0.11\linewidth]{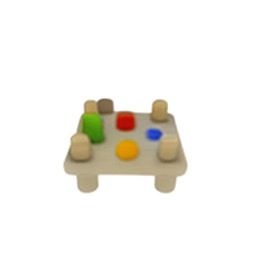} &
\includegraphics[trim=40 40 40 40, clip, width=0.11\linewidth]{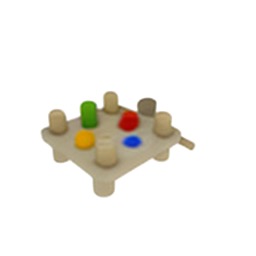} &
\includegraphics[trim=40 40 40 40, clip, width=0.11\linewidth]{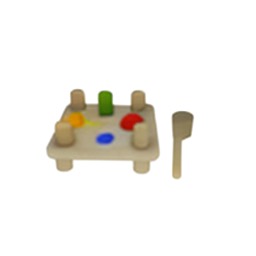} &
\includegraphics[trim=40 40 40 40, clip, width=0.11\linewidth]{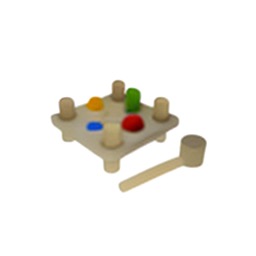} &
\includegraphics[trim=40 40 40 40, clip, width=0.11\linewidth]{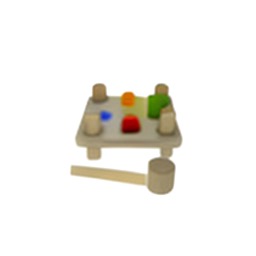} \\

\includegraphics[trim=50 50 50 50, clip, width=0.11\linewidth]{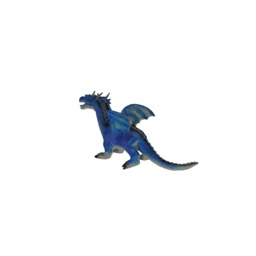} &
\includegraphics[trim=50 50 50 50, clip, width=0.11\linewidth]{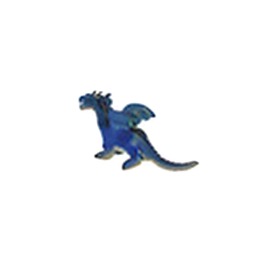} &
\includegraphics[trim=50 50 50 50, clip, width=0.11\linewidth]{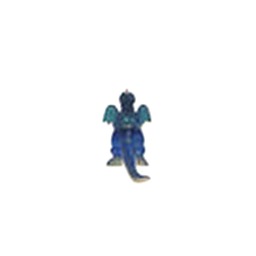} &
\includegraphics[trim=50 50 50 50, clip, width=0.11\linewidth]{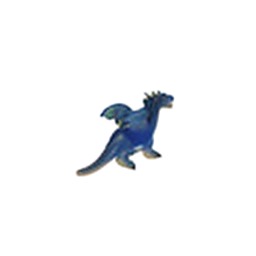} &
\includegraphics[trim=50 50 50 50, clip, width=0.11\linewidth]{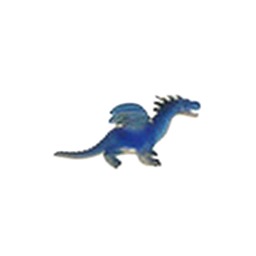} &
\includegraphics[trim=50 50 50 50, clip, width=0.11\linewidth]{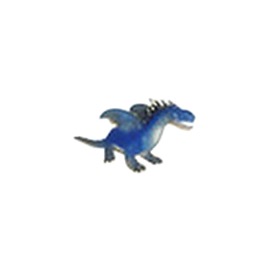} &
\includegraphics[trim=50 50 50 50, clip, width=0.11\linewidth]{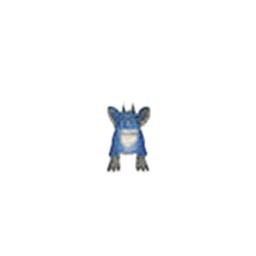} &
\includegraphics[trim=50 50 50 50, clip, width=0.11\linewidth]{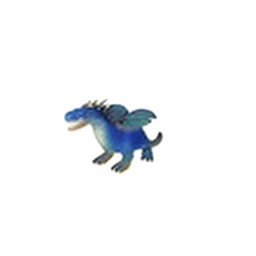} &
\includegraphics[trim=50 50 50 50, clip, width=0.11\linewidth]{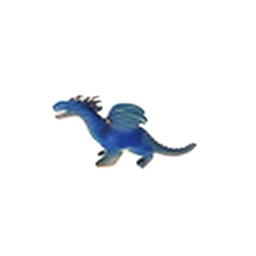} \\

Input image & \multicolumn{8}{c}{3D consistent images generated by \methodname + Zero123~\cite{zero123} }

\end{tabular}
\end{center}
\vspace{-0.6cm}
\captionof{figure}{\textbf{Google Scan Objects dataset.} More qualitative results generated by our method. }\label{fig:more_vis}
\vspace{-0.9cm}
\end{figure*}

\begin{figure*}[t]
\vspace{-0.6cm}
\begin{center}
\setlength\tabcolsep{1pt}
\small
\resizebox{\linewidth}{!}{
\begin{tabular}{m{0.09\linewidth}|c|m{0.09\linewidth}m{0.09\linewidth}m{0.09\linewidth}m{0.09\linewidth}m{0.09\linewidth}m{0.09\linewidth}m{0.09\linewidth}m{0.09\linewidth}}   

&
\specialcell[c]{Zero123\cite{zero123} \\ (5 sec)} 

&
\includegraphics[trim=30 30 30 30, clip, width=\linewidth]{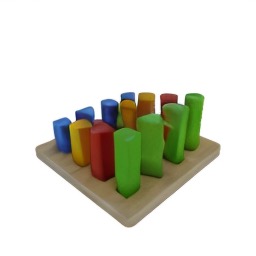} &
\includegraphics[trim=30 30 30 30, clip, width=\linewidth]{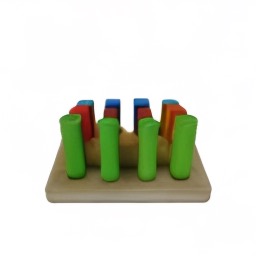} &
\includegraphics[trim=30 30 30 30, clip, width=\linewidth]{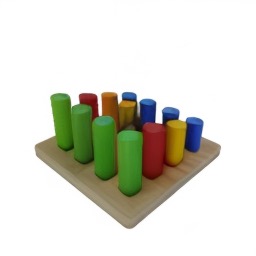} &
\includegraphics[trim=30 30 30 30, clip, width=\linewidth]{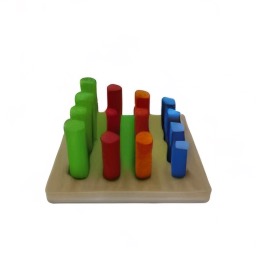} &
\includegraphics[trim=30 30 30 30, clip, width=\linewidth]{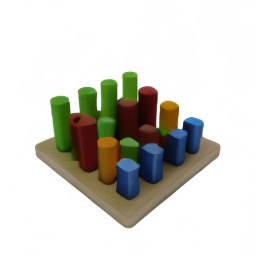} &
\includegraphics[trim=30 30 30 30, clip, width=\linewidth]{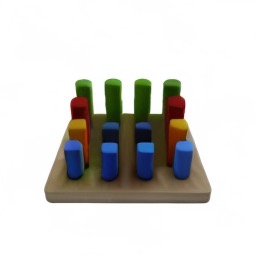} &
\includegraphics[trim=30 30 30 30, clip, width=\linewidth]{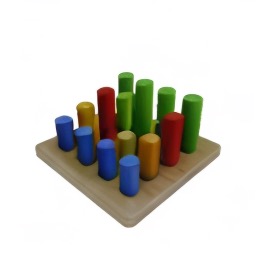} &
\includegraphics[trim=30 30 30 30, clip, width=\linewidth]{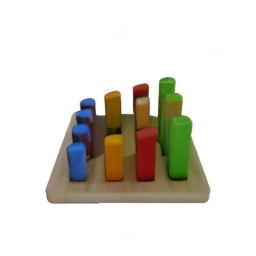} \\

& \specialcell[c]{DreamFusion\cite{poole2022dreamfusion}+\\Zero123\cite{poole2022dreamfusion,zero123} \\ (18 min)\\~} &
\includegraphics[trim=30 30 30 30, clip, width=\linewidth]{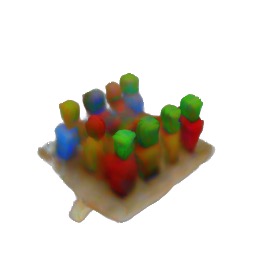} &
\includegraphics[trim=30 30 30 30, clip, width=\linewidth]{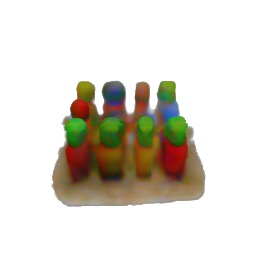} &
\includegraphics[trim=30 30 30 30, clip, width=\linewidth]{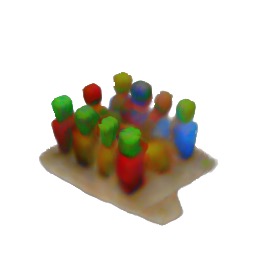} &
\includegraphics[trim=30 30 30 30, clip, width=\linewidth]{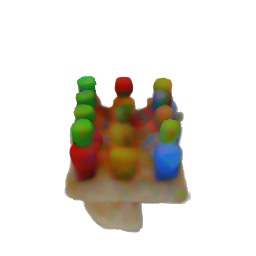} &
\includegraphics[trim=30 30 30 30, clip, width=\linewidth]{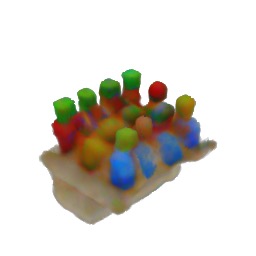} &
\includegraphics[trim=30 30 30 30, clip, width=\linewidth]{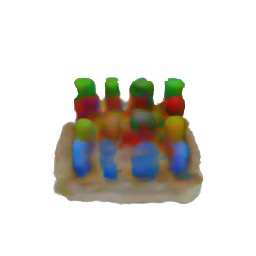} &
\includegraphics[trim=30 30 30 30, clip, width=\linewidth]{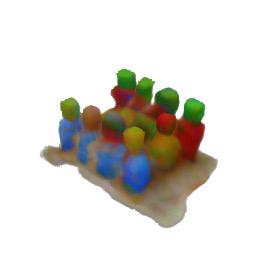} &
\includegraphics[trim=30 30 30 30, clip, width=\linewidth]{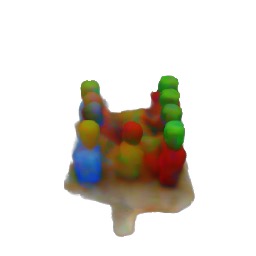} \\

\includegraphics[trim=10 10 10 10, clip, width=\linewidth]{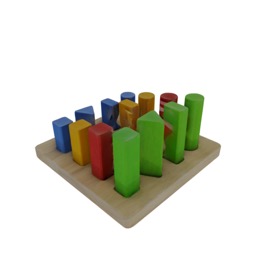} & \specialcell[c]{Syncdreamer\cite{liu2023syncdreamer} \\ (2 min)} &
\includegraphics[trim=30 30 30 30, clip, width=\linewidth]{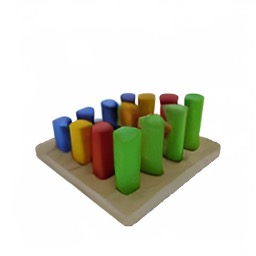} &
\includegraphics[trim=30 30 30 30, clip, width=\linewidth]{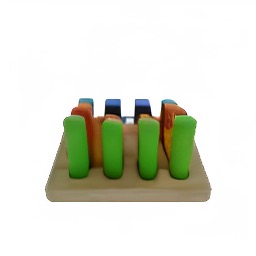} &
\includegraphics[trim=30 30 30 30, clip, width=\linewidth]{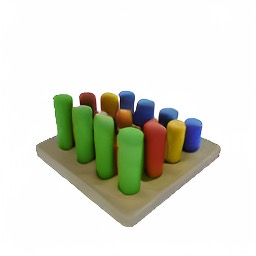} &
\includegraphics[trim=30 30 30 30, clip, width=\linewidth]{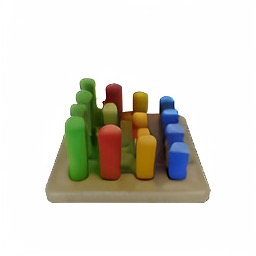} &
\includegraphics[trim=30 30 30 30, clip, width=\linewidth]{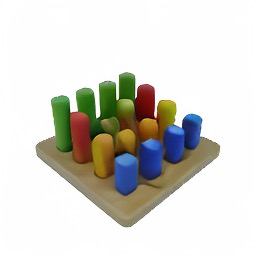} &
\includegraphics[trim=30 30 30 30, clip, width=\linewidth]{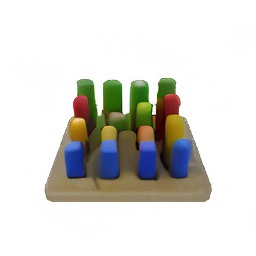} &
\includegraphics[trim=30 30 30 30, clip, width=\linewidth]{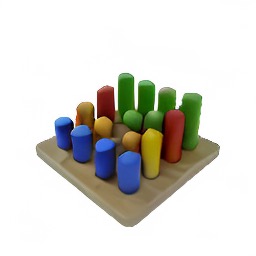} &
\includegraphics[trim=30 30 30 30, clip, width=\linewidth]{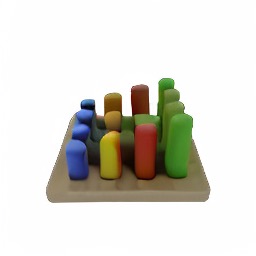} \\

& \specialcell[c]{Ours \\ (40 sec)} &
\includegraphics[trim=30 30 30 30, clip, width=\linewidth]{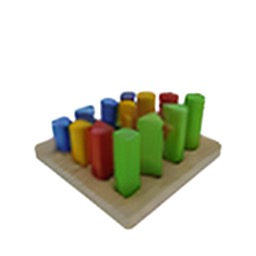} &
\includegraphics[trim=30 30 30 30, clip, width=\linewidth]{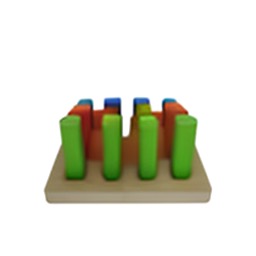} &
\includegraphics[trim=30 30 30 30, clip, width=\linewidth]{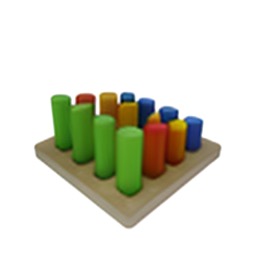} &
\includegraphics[trim=30 30 30 30, clip, width=\linewidth]{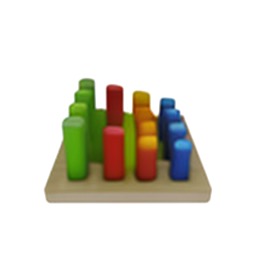} &
\includegraphics[trim=30 30 30 30, clip, width=\linewidth]{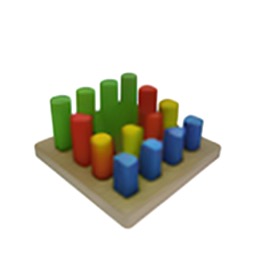} &
\includegraphics[trim=30 30 30 30, clip, width=\linewidth]{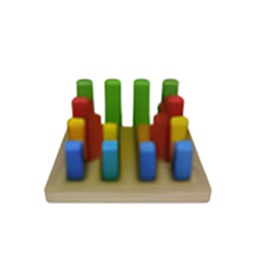} &
\includegraphics[trim=30 30 30 30, clip, width=\linewidth]{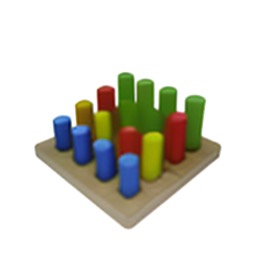} &
\includegraphics[trim=30 30 30 30, clip, width=\linewidth]{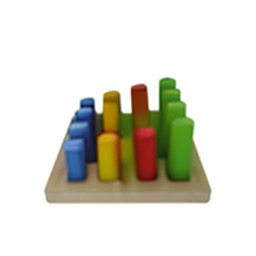} \\

& Ground truth &
\includegraphics[trim=30 30 30 30, clip, width=\linewidth]{Samples/peg/gt_0.jpg} &
\includegraphics[trim=30 30 30 30, clip, width=\linewidth]{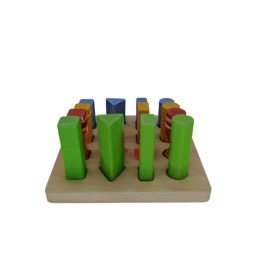} &
\includegraphics[trim=30 30 30 30, clip, width=\linewidth]{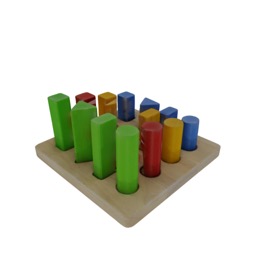} &
\includegraphics[trim=30 30 30 30, clip, width=\linewidth]{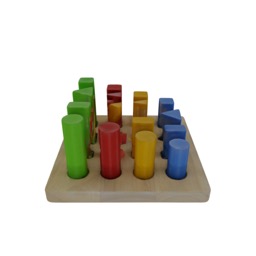} &
\includegraphics[trim=30 30 30 30, clip, width=\linewidth]{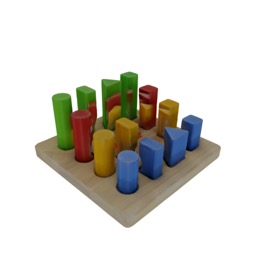} &
\includegraphics[trim=30 30 30 30, clip, width=\linewidth]{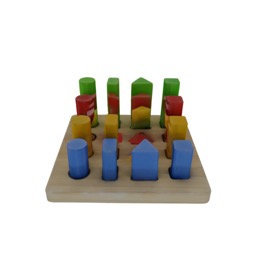} &
\includegraphics[trim=30 30 30 30, clip, width=\linewidth]{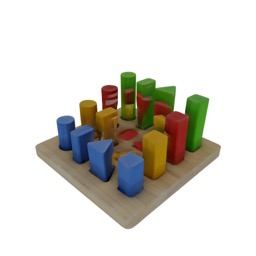} &
\includegraphics[trim=30 30 30 30, clip, width=\linewidth]{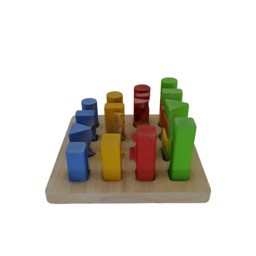} \\

\hline

 &
\specialcell[c]{Zero123\cite{zero123} \\ (5 sec)} &
\includegraphics[trim=40 40 40 40, clip, width=\linewidth]{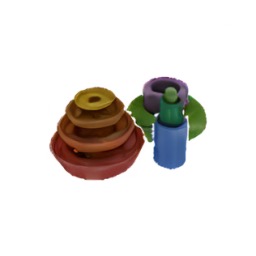} &
\includegraphics[trim=40 40 40 40, clip, width=\linewidth]{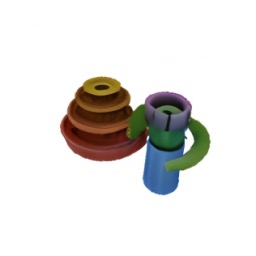} &
\includegraphics[trim=40 40 40 40, clip, width=\linewidth]{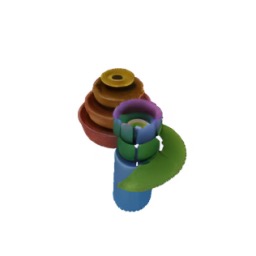} &
\includegraphics[trim=40 40 40 40, clip, width=\linewidth]{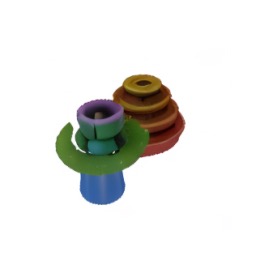} &
\includegraphics[trim=40 40 40 40, clip, width=\linewidth]{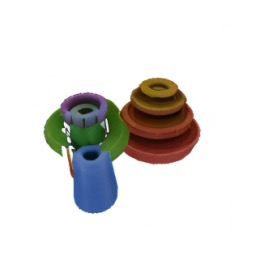} &
\includegraphics[trim=40 40 40 40, clip, width=\linewidth]{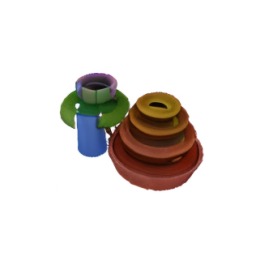} &
\includegraphics[trim=40 40 40 40, clip, width=\linewidth]{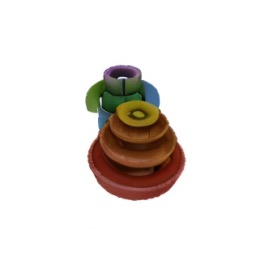} &
\includegraphics[trim=40 40 40 40, clip, width=\linewidth]{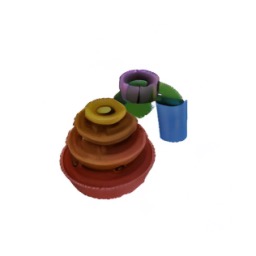} \\

& \specialcell[c]{DreamFusion\cite{poole2022dreamfusion}+\\Zero123\cite{poole2022dreamfusion,zero123} \\ (18 min)} &
\includegraphics[trim=60 60 60 60, clip, width=\linewidth]{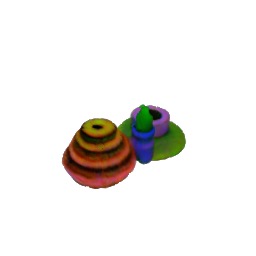} &
\includegraphics[trim=60 60 60 60, clip, width=\linewidth]{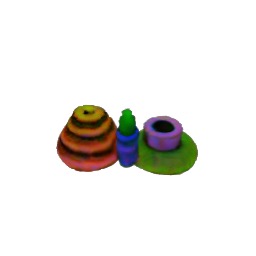} &
\includegraphics[trim=60 60 60 60, clip, width=\linewidth]{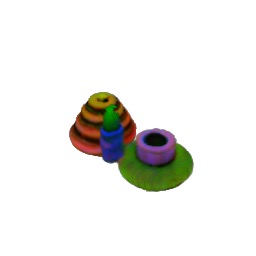} &
\includegraphics[trim=60 60 60 60, clip, width=\linewidth]{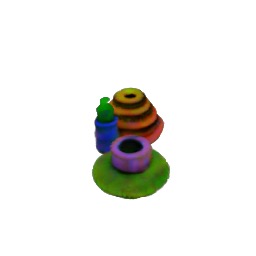} &
\includegraphics[trim=60 60 60 60, clip, width=\linewidth]{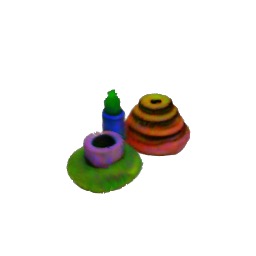} &
\includegraphics[trim=60 60 60 60, clip, width=\linewidth]{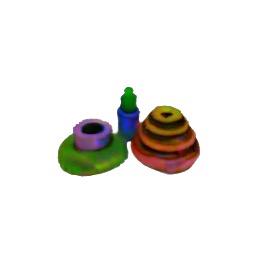} &
\includegraphics[trim=60 60 60 60, clip, width=\linewidth]{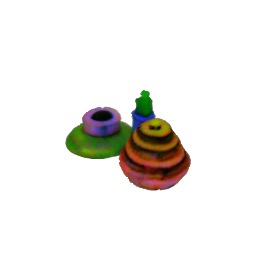} &
\includegraphics[trim=60 60 60 60, clip, width=\linewidth]{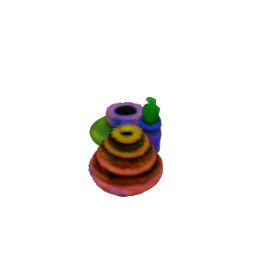} \\

\includegraphics[trim=60 60 60 60, clip, width=\linewidth]{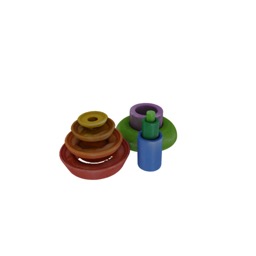} & \specialcell[c]{Syncdreamer\cite{liu2023syncdreamer} \\ (2 min)} &
\includegraphics[trim=60 60 60 60, clip, width=\linewidth]{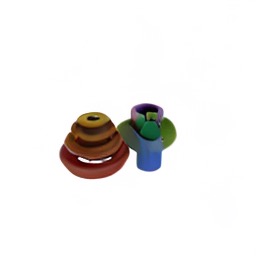} &
\includegraphics[trim=60 60 60 60, clip, width=\linewidth]{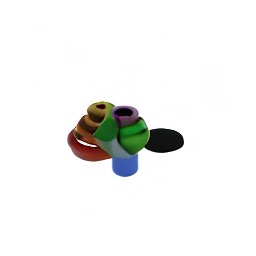} &
\includegraphics[trim=60 60 60 60, clip, width=\linewidth]{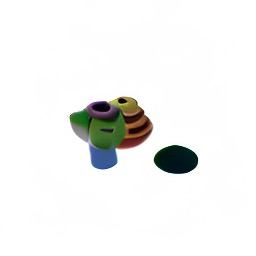} &
\includegraphics[trim=60 60 60 60, clip, width=\linewidth]{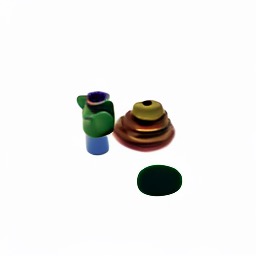} &
\includegraphics[trim=60 60 60 60, clip, width=\linewidth]{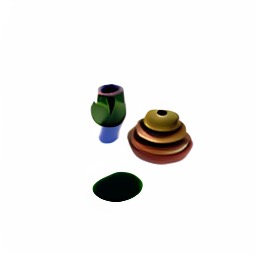} &
\includegraphics[trim=60 60 60 60, clip, width=\linewidth]{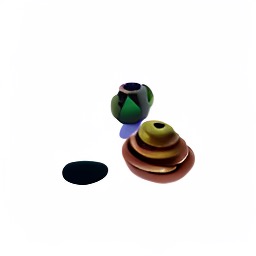} &
\includegraphics[trim=60 60 60 60, clip, width=\linewidth]{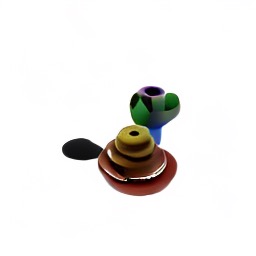} &
\includegraphics[trim=60 60 60 60, clip, width=\linewidth]{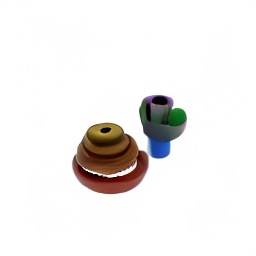} \\

& \specialcell[c]{Ours \\ (40 sec)} &
\includegraphics[trim=60 60 60 60, clip, width=\linewidth]{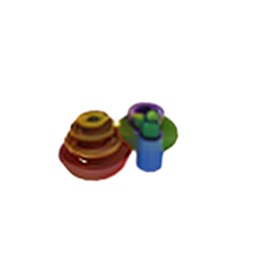} &
\includegraphics[trim=60 60 60 60, clip, width=\linewidth]{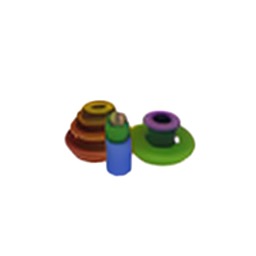} &
\includegraphics[trim=60 60 60 60, clip, width=\linewidth]{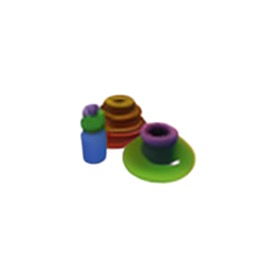} &
\includegraphics[trim=60 60 60 60, clip, width=\linewidth]{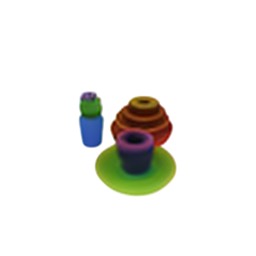} &
\includegraphics[trim=60 60 60 60, clip, width=\linewidth]{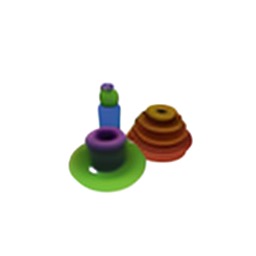} &
\includegraphics[trim=60 60 60 60, clip, width=\linewidth]{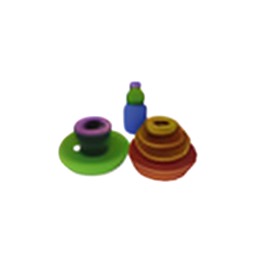} &
\includegraphics[trim=60 60 60 60, clip, width=\linewidth]{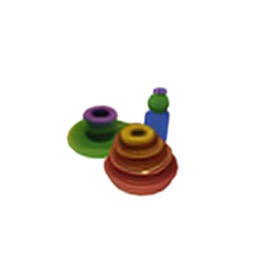} &
\includegraphics[trim=60 60 60 60, clip, width=\linewidth]{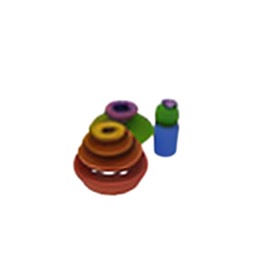} \\

& Ground truth &
\includegraphics[trim=60 60 60 60, clip, width=\linewidth]{Samples/cone/gt_0.jpg} &
\includegraphics[trim=60 60 60 60, clip, width=\linewidth]{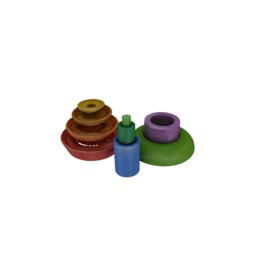} &
\includegraphics[trim=60 60 60 60, clip, width=\linewidth]{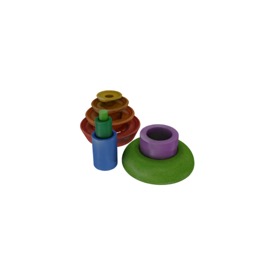} &
\includegraphics[trim=60 60 60 60, clip, width=\linewidth]{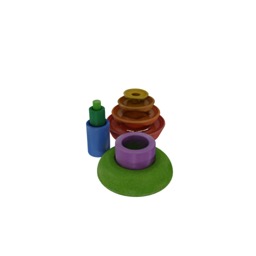} &
\includegraphics[trim=60 60 60 60, clip, width=\linewidth]{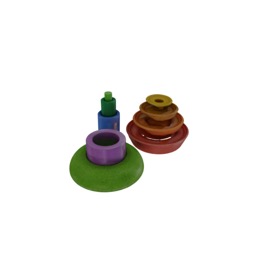} &
\includegraphics[trim=60 60 60 60, clip, width=\linewidth]{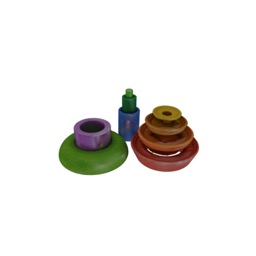} &
\includegraphics[trim=60 60 60 60, clip, width=\linewidth]{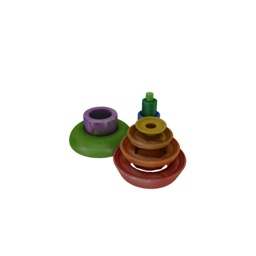} &
\includegraphics[trim=60 60 60 60, clip, width=\linewidth]{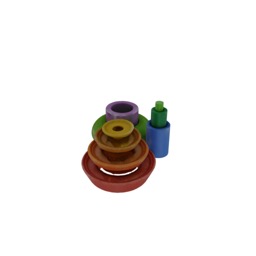} \\

\hline

Input image & Method &  \multicolumn{8}{c}{Generated Multi-view images}

\end{tabular}
}
\end{center}
\vspace{-0.4cm}
\captionof{figure}{\textbf{Google Scan Objects dataset.} Comparison of baseline methods and our approach with complex geometry and colors. Our method is able to infer the correct geometry and produce consistent multi-view images.}\label{fig:qualitative}
\vspace{-0.4cm}
\end{figure*}

\begin{figure*}[t]
\vspace{-0.2cm}
\begin{center}
\setlength\tabcolsep{1pt}
\small
\begin{tabular}{m{0.09\linewidth}|c|m{0.09\linewidth}m{0.09\linewidth}m{0.09\linewidth}m{0.09\linewidth}m{0.09\linewidth}m{0.09\linewidth}m{0.09\linewidth}m{0.09\linewidth}}

 &

Ground truth &
\includegraphics[trim=60 60 60 60, clip, width=\linewidth]{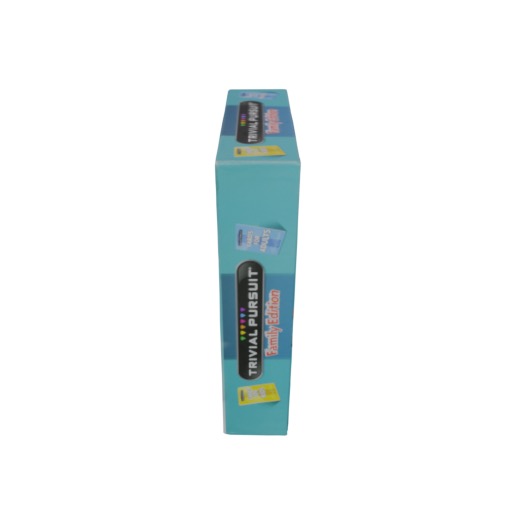} &
\includegraphics[trim=60 60 60 60, clip, width=\linewidth]{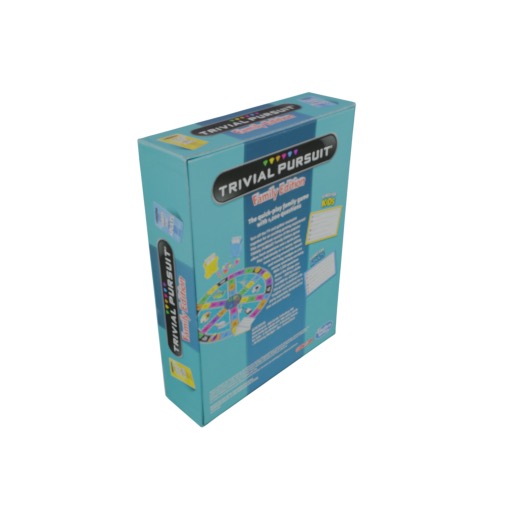} &
\includegraphics[trim=60 60 60 60, clip, width=\linewidth]{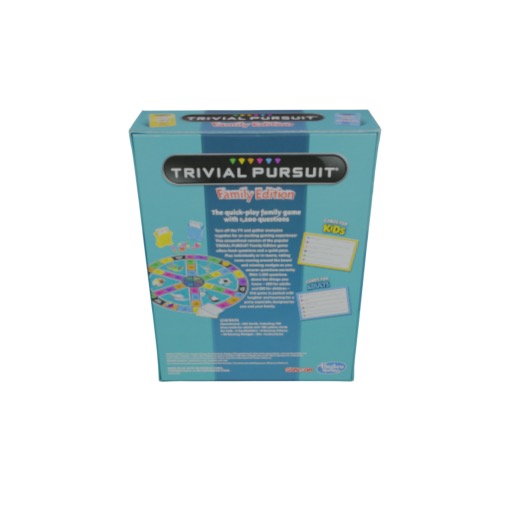} &
\includegraphics[trim=60 60 60 60, clip, width=\linewidth]{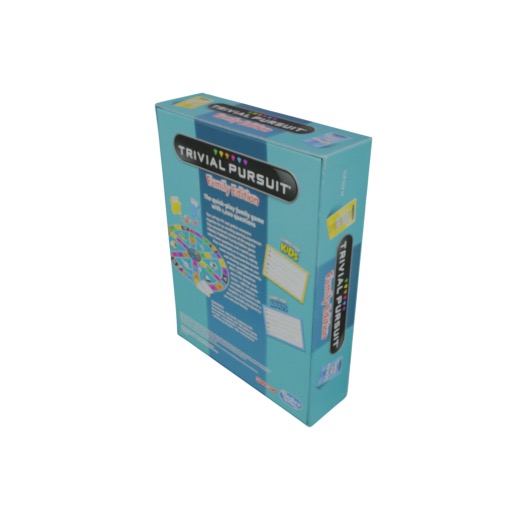} &
\includegraphics[trim=60 60 60 60, clip, width=\linewidth]{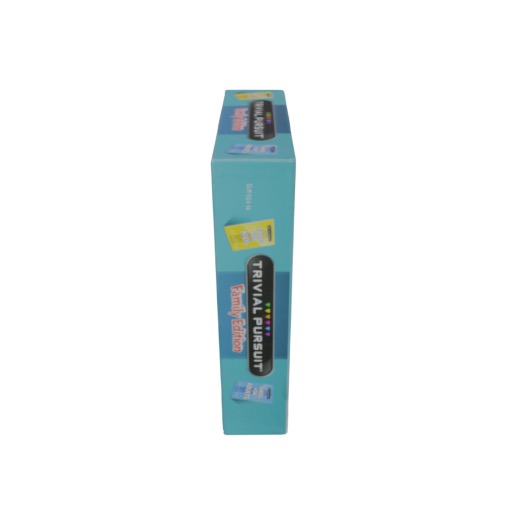} &
\includegraphics[trim=60 60 60 60, clip, width=\linewidth]{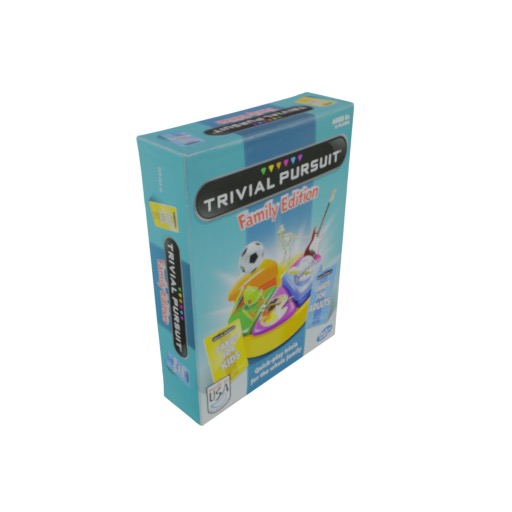} &
\includegraphics[trim=60 60 60 60, clip, width=\linewidth]{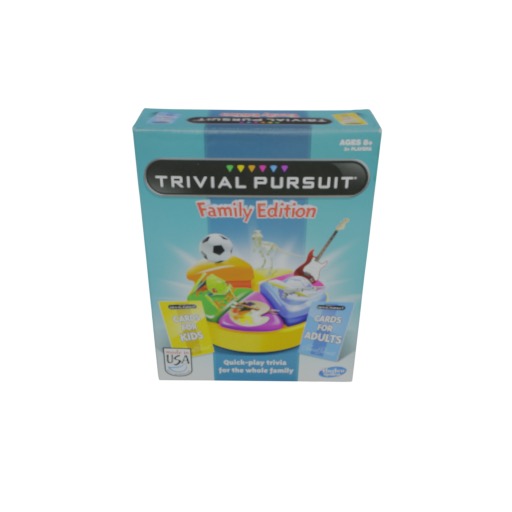} &
\includegraphics[trim=60 60 60 60, clip, width=\linewidth]{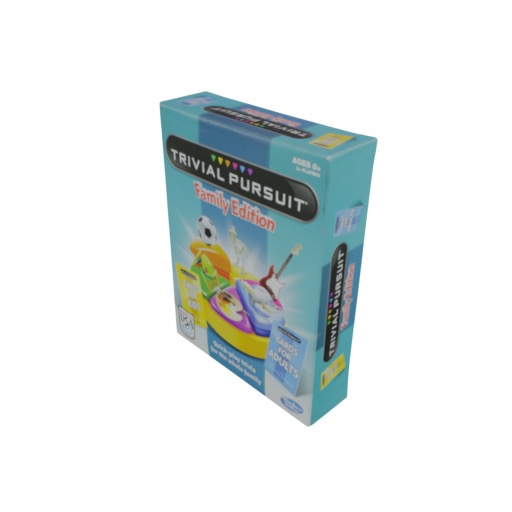} \\


\includegraphics[trim=60 60 60 60, clip, width=\linewidth]{Samples/Hasbro_Trivial_Pursuit_Family_Edition_Game/gt/bg_00_59.jpg} & \specialcell[c]{Ours \\Sample 1} &
\includegraphics[trim=25 25 25 25, clip, width=\linewidth]{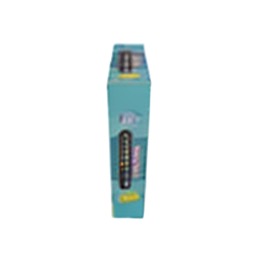} &
\includegraphics[trim=25 25 25 25, clip, width=\linewidth]{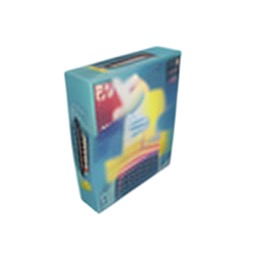} &
\includegraphics[trim=25 25 25 25, clip, width=\linewidth]{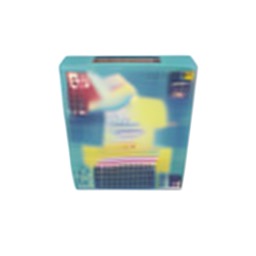} &
\includegraphics[trim=25 25 25 25, clip, width=\linewidth]{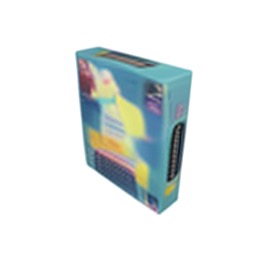} &
\includegraphics[trim=25 25 25 25, clip, width=\linewidth]{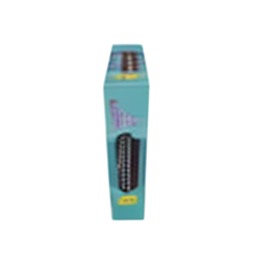} &
\includegraphics[trim=25 25 25 25, clip, width=\linewidth]{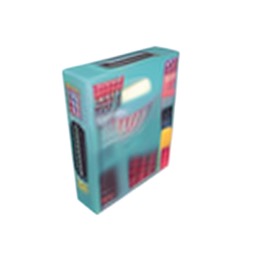} &
\includegraphics[trim=25 25 25 25, clip, width=\linewidth]{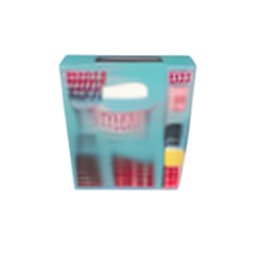} &
\includegraphics[trim=25 25 25 25, clip, width=\linewidth]{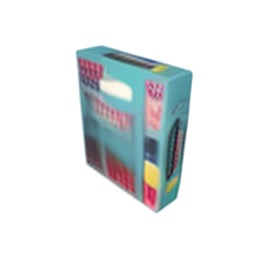} \\

& \specialcell[c]{Ours \\Sample 2} &
\includegraphics[trim=25 25 25 25, clip, width=\linewidth]{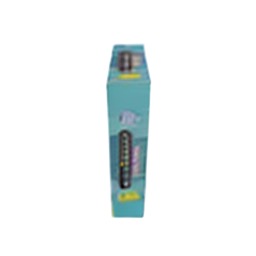} &
\includegraphics[trim=25 25 25 25, clip, width=\linewidth]{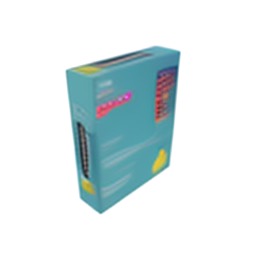} &
\includegraphics[trim=25 25 25 25, clip, width=\linewidth]{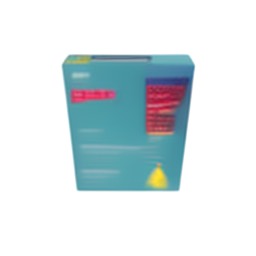} &
\includegraphics[trim=25 25 25 25, clip, width=\linewidth]{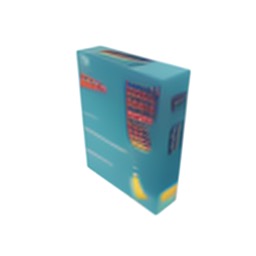} &
\includegraphics[trim=25 25 25 25, clip, width=\linewidth]{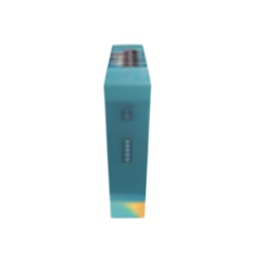} &
\includegraphics[trim=25 25 25 25, clip, width=\linewidth]{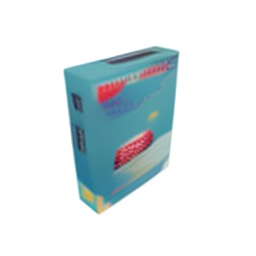} &
\includegraphics[trim=25 25 25 25, clip, width=\linewidth]{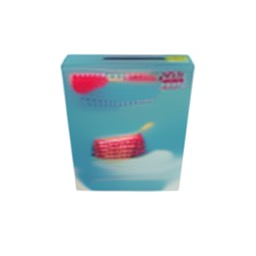} &
\includegraphics[trim=25 25 25 25, clip, width=\linewidth]{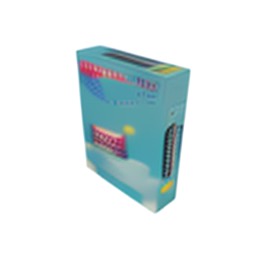} \\
\hline

 &

Ground truth &

\includegraphics[trim=60 60 60 60, clip, width=\linewidth]{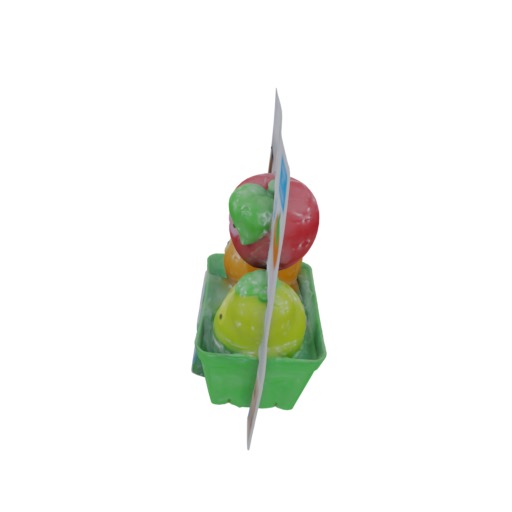} &
\includegraphics[trim=60 60 60 60, clip, width=\linewidth]{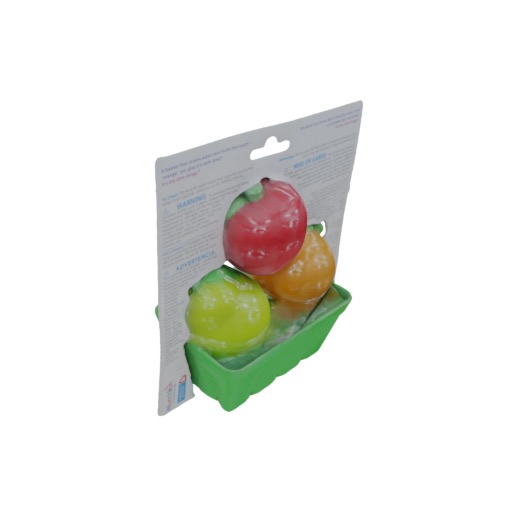} &
\includegraphics[trim=60 60 60 60, clip, width=\linewidth]{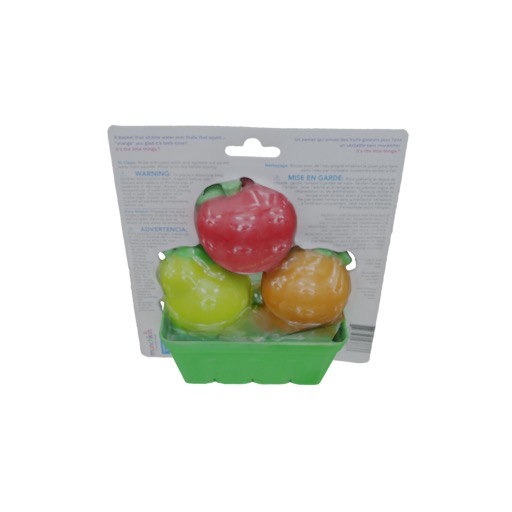} &
\includegraphics[trim=60 60 60 60, clip, width=\linewidth]{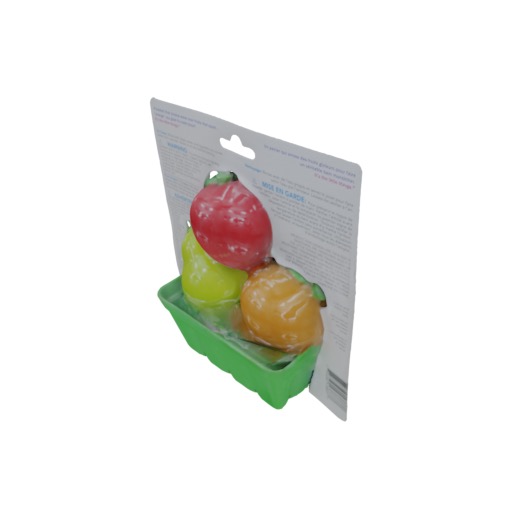} &
\includegraphics[trim=60 60 60 60, clip, width=\linewidth]{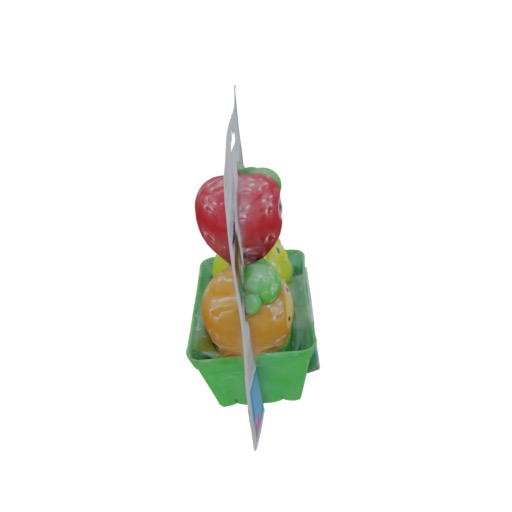} &
\includegraphics[trim=60 60 60 60, clip, width=\linewidth]{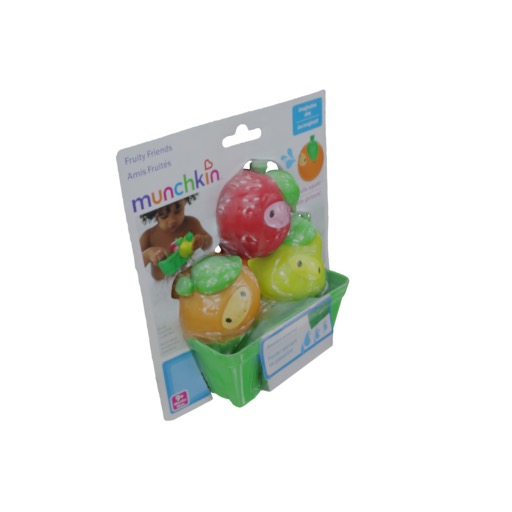} &
\includegraphics[trim=60 60 60 60, clip, width=\linewidth]{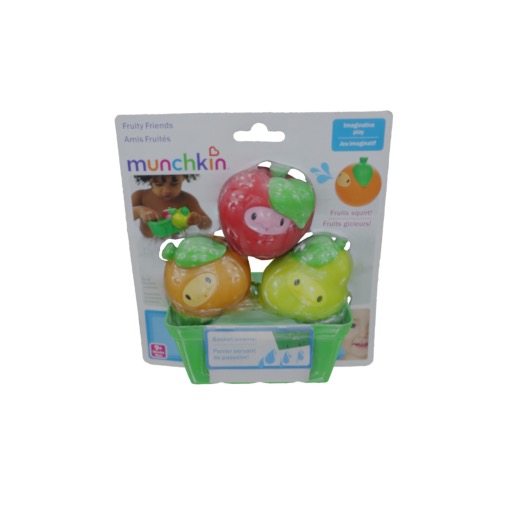} &
\includegraphics[trim=60 60 60 60, clip, width=\linewidth]{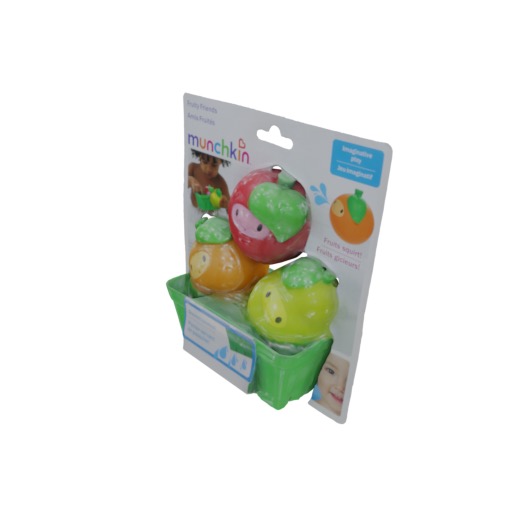} \\

\includegraphics[trim=60 60 60 60, clip, width=\linewidth]{Samples/Fruity_Friends/gt/bg_00_59.jpg} & 
\specialcell[c]{Ours \\Sample 1} &
\includegraphics[trim=25 25 25 25, clip, width=\linewidth]{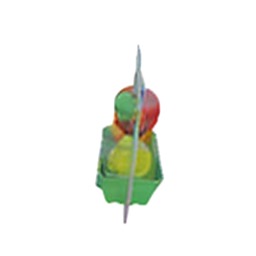} &
\includegraphics[trim=25 25 25 25, clip, width=\linewidth]{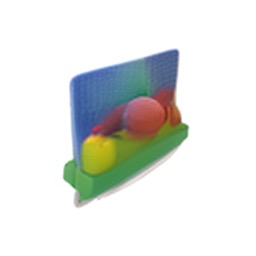} &
\includegraphics[trim=25 25 25 25, clip, width=\linewidth]{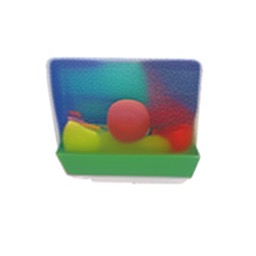} &
\includegraphics[trim=25 25 25 25, clip, width=\linewidth]{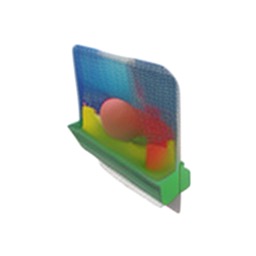} &
\includegraphics[trim=25 25 25 25, clip, width=\linewidth]{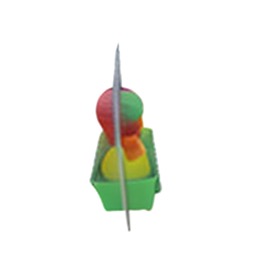} &
\includegraphics[trim=25 25 25 25, clip, width=\linewidth]{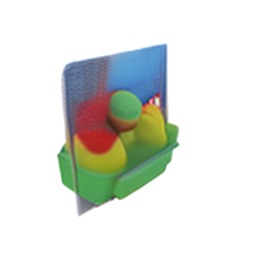} &
\includegraphics[trim=25 25 25 25, clip, width=\linewidth]{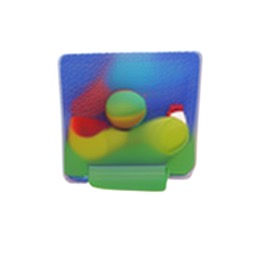} &
\includegraphics[trim=25 25 25 25, clip, width=\linewidth]{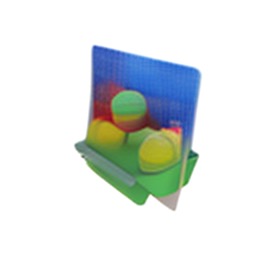} \\

 & 
\specialcell[c]{Ours \\Sample 2} &
\includegraphics[trim=25 25 25 25, clip, width=\linewidth]{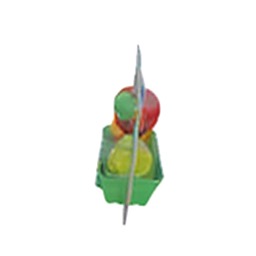} &
\includegraphics[trim=25 25 25 25, clip, width=\linewidth]{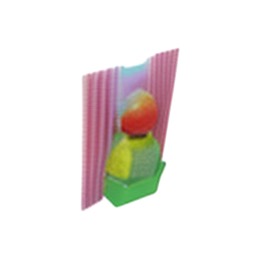} &
\includegraphics[trim=25 25 25 25, clip, width=\linewidth]{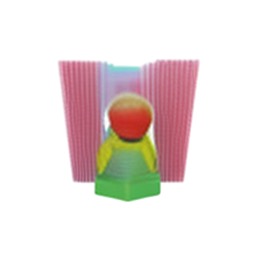} &
\includegraphics[trim=25 25 25 25, clip, width=\linewidth]{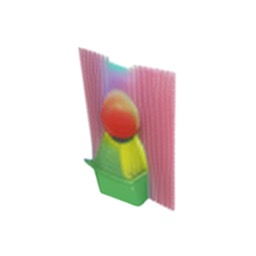} &
\includegraphics[trim=25 25 25 25, clip, width=\linewidth]{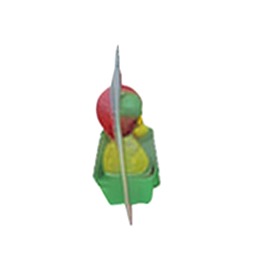} &
\includegraphics[trim=25 25 25 25, clip, width=\linewidth]{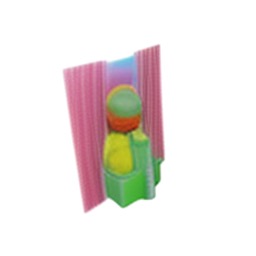} &
\includegraphics[trim=25 25 25 25, clip, width=\linewidth]{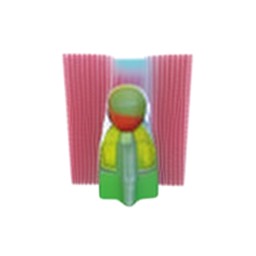} & 
\includegraphics[trim=25 25 25 25, clip, width=\linewidth]{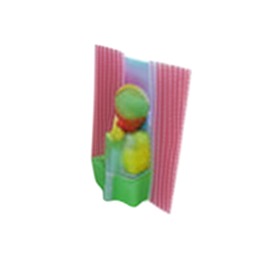} \\

\hline

Input image & Method &  \multicolumn{8}{c}{Generated Multi-view images}

\end{tabular}
\end{center}
\vspace{-0.4cm}
\captionof{figure}{\textbf{Google Scan Objects dataset.} Comparison of ground-truth multi-view images with two generated variants, showing the inherent ambiguity and diverse possibilities when extrapolating from a single reference image. }\label{fig:metric_problem}
\vspace{-0.3cm}
\end{figure*}

\subsection{Metrics}
We use following evaluation metrics to quantitatively evaluate the performance of our model compared to existing methods. Perceptual Loss (LPIPS)~\cite{zhang2018unreasonable} measures the perceptual distance between two images by comparing the deep features extracted by deep neural networks given each image as input. We use two pre-trained network AlexNet~\cite{krizhevsky2012imagenet} and VGG~\cite{simonyan2014very} to compute perceptual loss, denoted as LPIPS\_Alex and LPIPS\_VGG accordingly. Structural Similarity (SSIM)~\cite{wang2004image,wang2003multiscale} measures the structural similarity between two images considering both color and texture information. We report SSIM score~\cite{wang2004image} and multi-scale SSIM score~\cite{wang2003multiscale}, denoted as SSIM and MS-SSIM accordingly. We also report the Peak Signal-to-Noise Ratio (PSNR).

\subsection{Compared methods}
We compare our pipeline with several baseline methods. \zero serves as a baseline that is trained to produce multi-view images conditioned solely base on the CLIP embedding of input image and viewing angles. Additionally, we include the results from DreamFusion \cite{poole2022dreamfusion} using \zero as guidance. While DreamFusion \cite{poole2022dreamfusion} is multi-view consistent by construction, it requires training of a NeRF \cite{mildenhall2021nerf} through Score Distillation Sampling loss, resulting in long generation time. We also include a concurrent method, Syncdreamer \cite{liu2023syncdreamer}, in our comparison. We use its official inference code and provided pre-trained model for evaluation.

\subsection{Quantitative Results}
For each object in the Google Scanned Objects dataset, we use one of it's rendered image as reference image input to generate 16 uniformly sampled surrounding views of the object. We evaluate under three choices of elevation angle, 0 degree, 15 degree and 30 degree, to better reflect the model performance. Results are shown in Tab. \ref{tab:quantitative_results}. On elevation 0 and 15, our model largely outperform all existing works and concurrent work SyncDreamer~\cite{liu2023syncdreamer} on all metrics. On elevation 30, our model perform comparably to SyncDreamer. Comparing with our base model \zero~\cite{zero123}, plugging in our module improves its generation quality on all elevation angles. 

\subsection{Qualitative Results}
We show qualitative results generated by our method in Fig.~\ref{fig:title} and Fig.~\ref{fig:more_vis}. Our model can generalize well to unseen data. Moreover, we select two objects from Google Scanned Objects with complex geometry and diverse color to qualitatively compare with existing methods. Results are shown in Fig. \ref{fig:qualitative}. Our model improves 3D consistency of our base model \zero.

\noindent\textbf{Discussion.}
The task of generating unseen multi-view images of an object from a single reference image is severely ill-posed. In general, there are infinite numbers of possible solutions given only a single reference image.  Fig.~\ref{fig:metric_problem} illustrate such a solution ambiguity. Consequently, using so-called 'quantitative evaluation' by simply comparing the generated views with the ground-truth views as the performance metric is not well suited.  Designing better metrics for this task is an important future task.

\section{Conclusion}
We have proposed \methodname, a multi-view consistency plug-in block for latent diffusion models to improve 3D consistency without requiring explicit pixel correspondences or depth prediction. Experiments show that our models effectively improve 3D consistency of a frozen \zero backbone and can generalize well to unseen data. In the future we plan to further improve the computational efficiency of the model and develop a 3D reconstruction plug-in module to generate a 3D mesh along the multi-view image denoising process.

{\small
\bibliographystyle{ieee_fullname}
\bibliography{egbib}
}

\end{document}